\documentclass[]{fairmeta}
\usepackage{horizonlab}

\usepackage{bbding}
\usepackage{colortbl}
\usepackage{float}
\usepackage{ulem}
\usepackage{soul}
\usepackage{wrapfig}
\usepackage{array}
\usepackage{makecell}
\usepackage{utfsym}
\usepackage{cancel}
\usepackage{enumitem}
\usepackage{amsmath}
\usepackage{algorithmic}
\usepackage{xspace}

% ...

\newcommand{\logoleft}{}
\newcommand{\logoright}{}

\title{
CogWAM: Aligning Semantic Cognition with World Action Modeling
via Event-Driven Interfaces
}

\author[1,2,*]{\calmfont{Sen Wang}}
\author[2,\ddagger]{\calmfont{Liu Liu}}
\author[2]{\calmfont{Xinjiang Wang}}
\author[2]{\calmfont{Zequn Chen}}
\author[3]{\calmfont{Haoyi Jiang}}
\author[2]{\calmfont{Taojun Ding}}
\author[2]{\calmfont{Tingyang Xiao}}
\author[2,\dagger]{\calmfont{Zhizhong Su}}
\author[4]{\calmfont{Jie Wang}}
\author[1,\dagger]{\calmfont{Sanping Zhou}}

\affiliation[1]{\calmfont{Xi'an Jiaotong University}}

\affiliation[2]{\calmfont{Horizon Robotics}}

\affiliation[3]{\calmfont{
Huazhong University of Science and Technology
}}

\affiliation[4]{\calmfont{%
\unskip\nobreak\hspace{0.25em}University of Science and Technology of China
}}

\contribution[\ddagger]{Project leads}

\contribution[\dagger]{Corresponding authors}

\contribution[*]{Work done during an internship at Horizon Robotics.}

\metadata[Code]{
\url{https://github.com/HorizonRobotics/CogWAM}
}

\metadata[Webpage]{
\url{https://horizonrobotics.github.io/CogWAM/}
\hfill
\smash{
    \raisebox{-0.50cm}{
        \includegraphics[height=1.10cm]{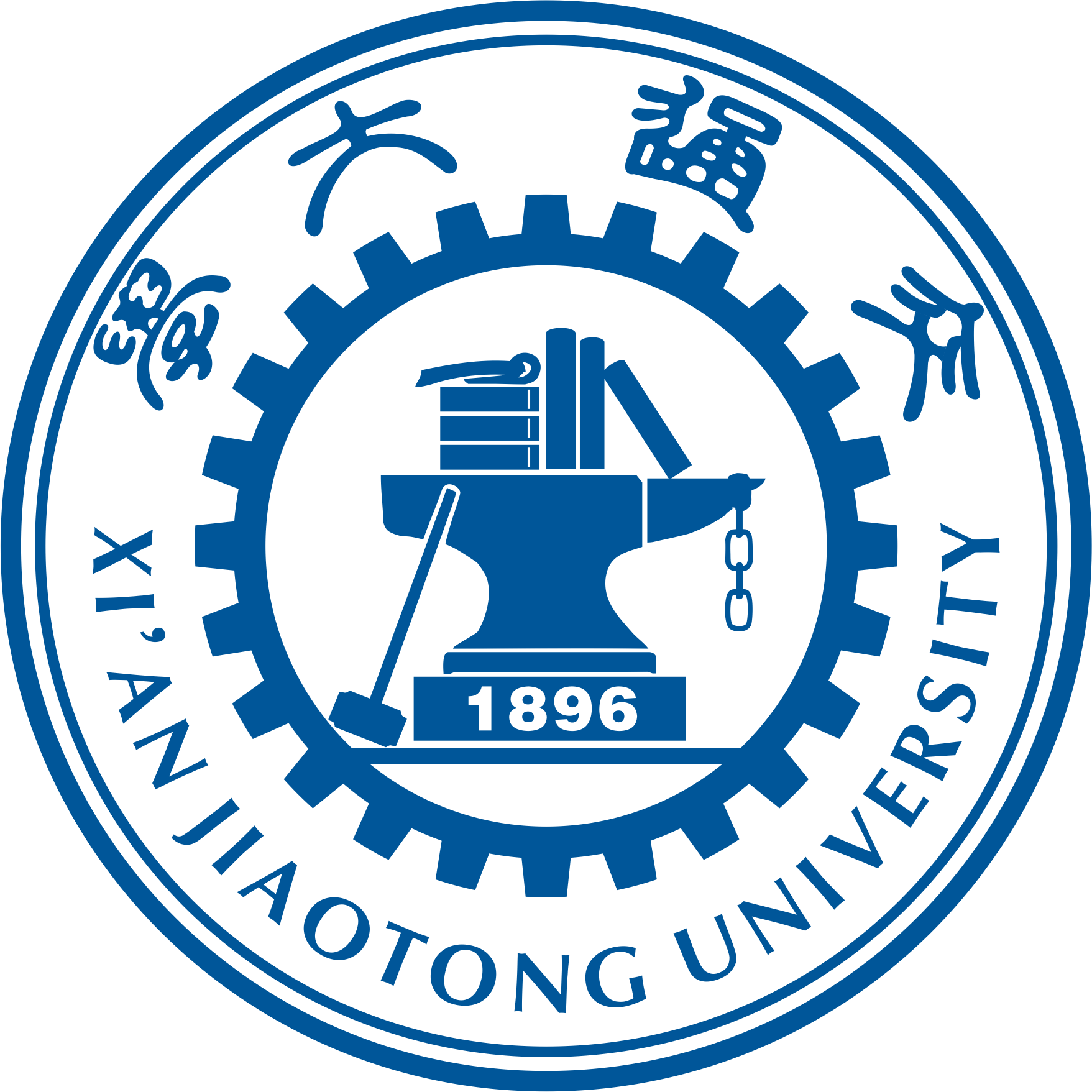}
        \hspace{0.06cm}
        \includegraphics[height=1.10cm]{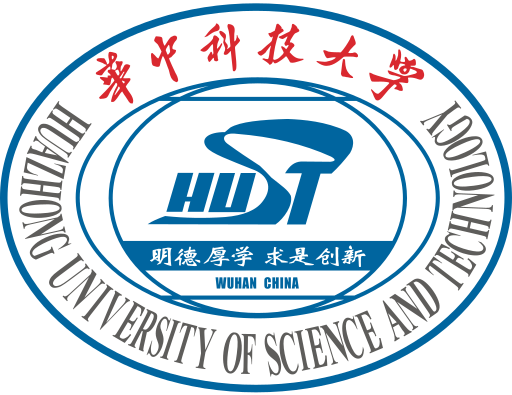}
        \hspace{0.06cm}
        \includegraphics[height=1.10cm]{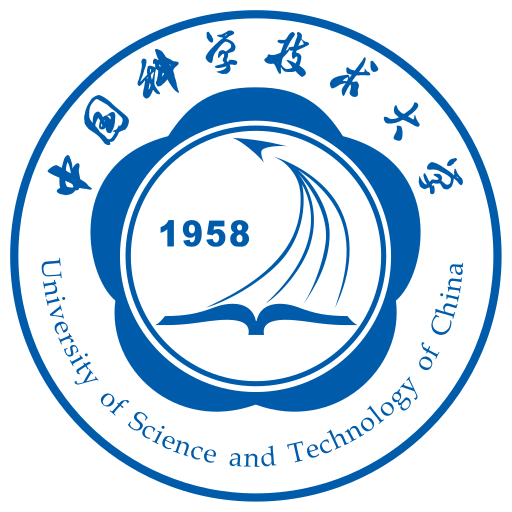}
        \hspace{0.06cm}
        \includegraphics[height=1.10cm]{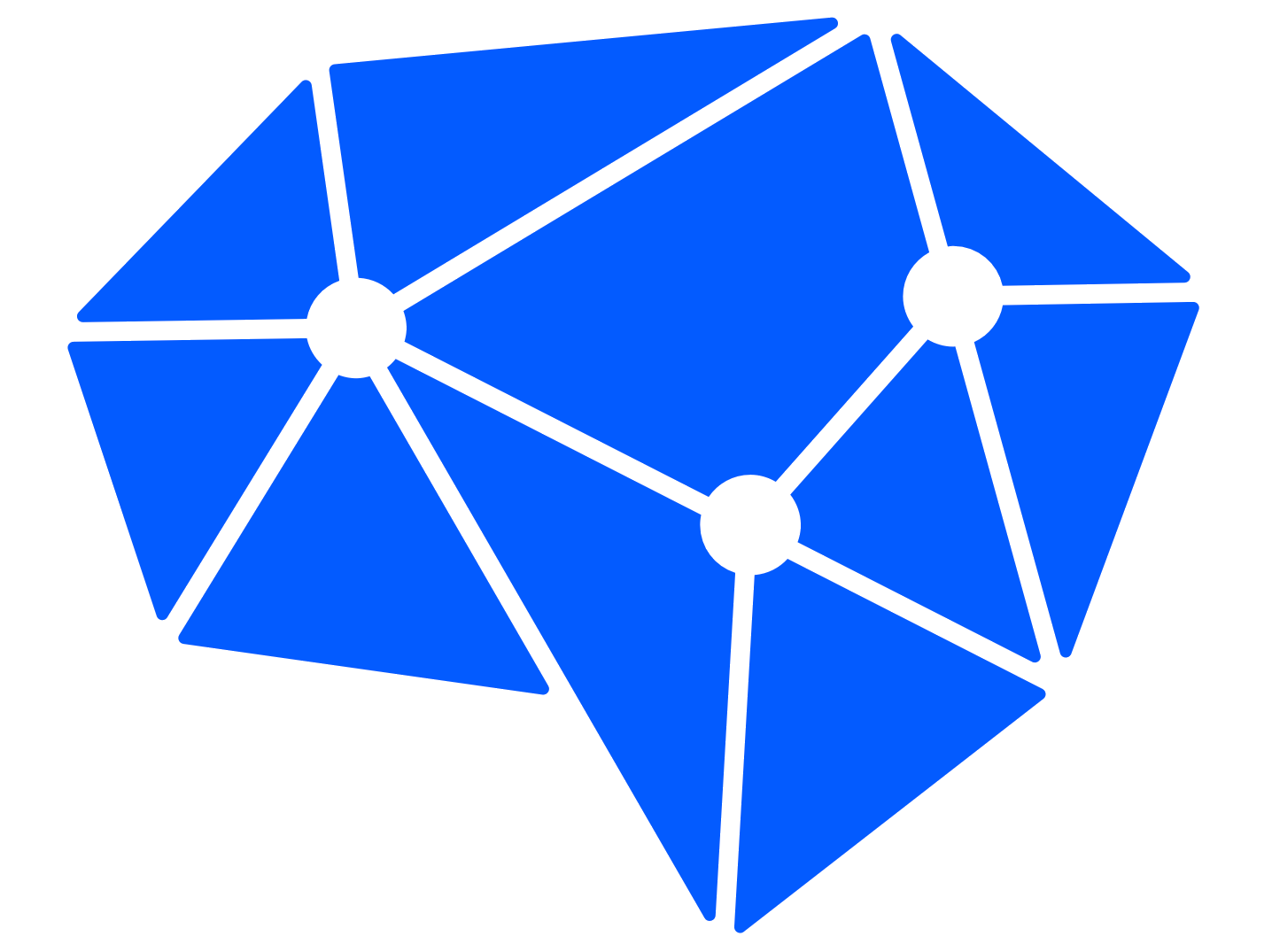}
    }
}
}
\metadata[Model]{
\url{https://huggingface.co/HorizonRobotics/CogWAM}
}

\begin{document}


%
%
% ...

\abstract{
Robot policies increasingly incorporate semantic reasoning and future-world prediction, yet combining these capabilities does not guarantee that local predictions and actions remain aligned with task progress. We introduce CogWAM, a cognition-guided world-action model that establishes an explicit semantic interface between task reasoning and world-action learning through a persistent Semantic State, which stores completed task events and the active subtask. CogWAM updates this state only when observations indicate semantic transitions, allowing task-level context to persist across multiple action chunks. To bridge semantic context with physical prediction and control, CogWAM employs progress-conditioned WORLD and ACTION queries that selectively extract task-relevant information for future-world prediction and action generation. During training, the Semantic State provides shared task-progress context for both branches, while inference removes the future-prediction branch and directly generates actions from observations and the maintained state. We further introduce semantic training strategies to improve transition learning and closed-loop conditioning. Without additional robot-action pretraining, CogWAM achieves 15.56 / 11.70\% Score/SR on RoboDojo and state-of-the-art performance on BiCoord, while real-world experiments demonstrate closed-loop dual-arm manipulation with 16.4× fewer Semantic State regenerations than step-wise updating. 

}


\maketitle


\section{Introduction}
\label{sec:introduction}

The promise of physical intelligence lies in embodied agents that can transform abstract intentions into purposeful changes in the physical world~\citep{ahn2022can,brooks1991intelligence,driess2023palm,intelligence2025pi_05}. Doing so requires more than predicting isolated actions. As interaction unfolds, a robot needs task-level cognition to relate what has already been accomplished to what should happen next, while grounding its evolving intent in physical interaction~\citep{huang2022inner,shi2025hirobot}. Robot policies should therefore connect task semantics, execution progress, and physical dynamics within a unified closed loop.

Current robot foundation models have advanced complementary parts of this loop. Vision-language-action models (VLAs) leverage large-scale vision-language pretraining to achieve strong semantic understanding and multi-task generalization~\citep{kim2025fine,black2024pi_0}. Recent approaches further introduce high-level reasoning, temporal memory, and subtask prediction to better track execution progress~\citep{shi2025hirobot,shi2025memoryvla,yang2025seqvla,wang2026dim,su2026worldscape_policy_2}. In parallel, world-action models (WAMs) augment action learning with predictions of future world representations, providing supervision for spatial change and interaction dynamics~\citep{zhu2025uwm,ye2026dreamzero,lyu2026lda,chen2026lawam,zhao2026sgwam,lingbot-va2026,zhang2026native}. Together, these directions address what should be accomplished next and how physical interaction may evolve. However, incorporating both capabilities does not by itself specify how task progress should be represented, maintained, and shared between them. A locally plausible scene prediction or action may still correspond to a subtask that has already been completed or has not yet been reached. This motivates a shared representation of task progress that both branches can condition on and that is revised in response to observed progress.

\begin{figure*}[t]
    \centering
    \includegraphics[width=1.0\linewidth]{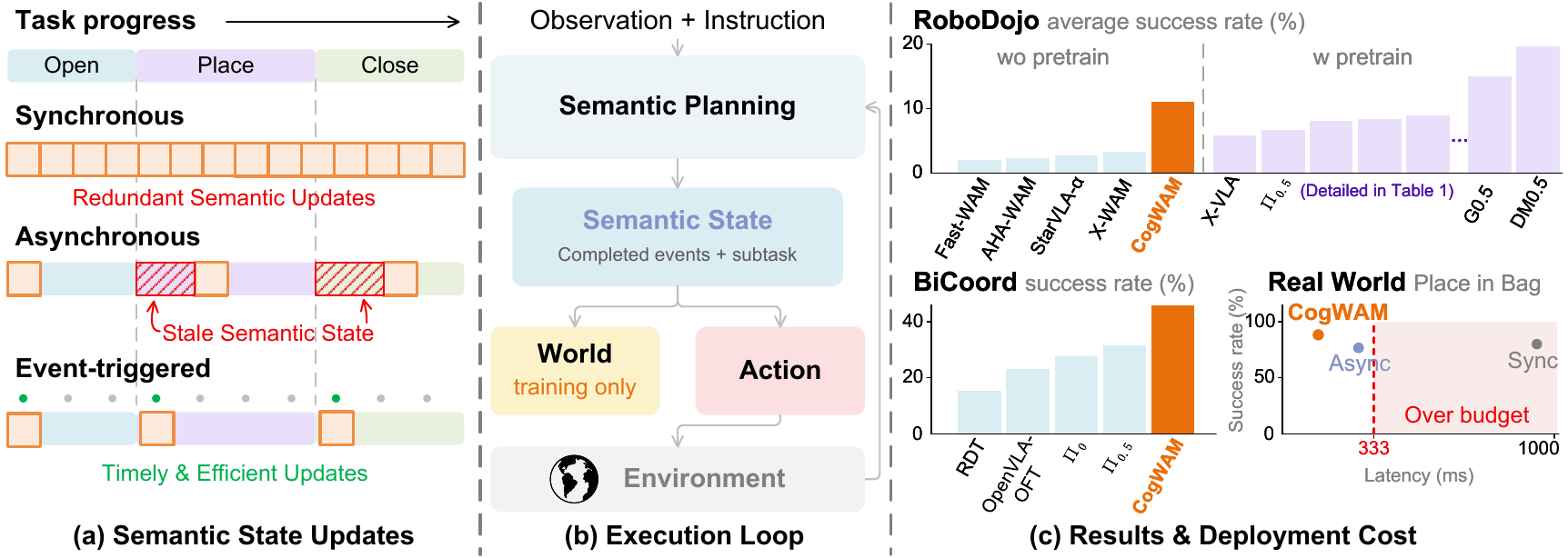}
    \caption{\textbf{Overview of CogWAM.}
    (a) Event-triggered updates maintain a persistent Semantic State while avoiding redundant updates and stale states.
    (b) The Semantic State couples semantic planning with world prediction and action generation through a closed-loop execution loop.
    (c) Benchmark results and real-world success--latency trade-off.}
    \label{fig:fig1}
\end{figure*}

To address this alignment problem, we introduce CogWAM (Fig.~\ref{fig:fig1}), which explicitly represents task progress as a persistent cognitive state rather than repeatedly inferring it from individual observations~\cite{yang2026world}. Existing hierarchical and memory-based policies demonstrate the importance of maintaining temporal context for long-horizon execution~\citep{shi2025hirobot,shi2025memoryvla,torne2026mem,yang2025seqvla}, yet they primarily focus on semantic reasoning or memory accumulation without explicitly coupling task progress with predictive world modeling and action generation. CogWAM addresses this gap through a persistent Semantic State that records completed task events and the active subtask. During execution, the state is updated only when observations indicate meaningful task transitions, allowing semantic context to persist across multiple action chunks. This event-driven formulation decouples the semantic timescale from the control timescale: actions require frequent replanning for reactive control~\citep{zhao2023learning}, whereas task semantics evolve only at a small number of stage transitions~\citep{torne2026mem,sutton1999between}.

Beyond maintaining task progress, an effective interface must also determine how semantic context is transferred to different physical objectives. Future-world prediction and action generation optimize different objectives, requiring representations that capture scene evolution and executable control. CogWAM therefore introduces progress-conditioned WORLD and ACTION queries, which are conditioned on the same Semantic State while selectively extracting branch-specific information for future multi-view latent prediction and continuous action generation. This allows task-level semantics to be shared while preserving objective-specific representations~\cite{li2023blip}.

This state-conditioned formulation introduces two training challenges. First, semantic transitions are sparse relative to state persistence, causing imbalance between UPDATE and KEEP decisions. Second, training with annotated states creates exposure bias because closed-loop execution relies on predicted states~\cite{bengio2015scheduled}. We address these issues with boundary-aware semantic supervision and progressive semantic conditioning.
Our main contributions are summarized as follows:
\begin{enumerate}[leftmargin=*, label=\arabic*., itemsep=0pt]
\item We present CogWAM, a semantics-aware world-action architecture that couples a persistent Semantic State with progress-conditioned WORLD/ACTION queries, enabling task-progress-aware future prediction and control.

\item We develop event-triggered semantic planning, which decouples semantic evolution from control frequency, together with training strategies that address sparse transitions and closed-loop semantic conditioning.

\item We demonstrate that explicit semantic interface design consistently improves robot policies across simulation, real-world, and different world-action architectures.
\end{enumerate}

\section{Related Work}

\textbf{Vision-Language-Action Models and Long-Horizon Reasoning.}
Vision-Language-Action (VLA) models leverage pretrained vision-language representations for language-conditioned robotic control~\citep{brohan2023rt,kim2024openvla,kim2025fine,zheng2026x}. Recent work increasingly augments such policies with higher-level reasoning and temporal context. $\pi_{0.5}$ incorporates high-level semantic prediction to support open-world and long-horizon manipulation~\citep{intelligence2025pi_05}, while Hi Robot explicitly decomposes high-level semantic reasoning and low-level visuomotor execution~\citep{shi2025hirobot}. MemoryVLA introduces perceptual--cognitive memory for temporally dependent control~\citep{shi2025memoryvla}, and Mem maintains multi-scale embodied memory across an episode~\citep{torne2026mem}. SeqVLA explicitly predicts subtask completion to control transitions during sequential manipulation~\citep{yang2025seqvla}. 
These works demonstrate the importance of reasoning beyond the current observation and global instruction. CogWAM similarly models execution progress, but differs by explicitly transferring maintained task progress into world-action learning through a structured semantic interface.

\textbf{World-Action Models and Predictive Robot Policies.}
WAMs augment action learning with predictive models of future scene evolution~\cite{du2023learning}. Unified World Models jointly model future observations and actions~\citep{zhu2025uwm}, while DreamZero scales this paradigm and demonstrates strong zero-shot physical generalization~\citep{ye2026dreamzero}. Recent studies increasingly question whether dense future video generation is necessary. Fast-WAM shows that world-model co-training can provide substantial benefits even without explicit future generation at inference time~\citep{yuan2026fast}. In parallel, recent approaches move toward compact, action-relevant future representations: LaWAM predicts latent visual subgoals that condition robot actions~\citep{chen2026lawam}, whereas SG-WAM models future dynamics directly in a geometry-aware policy representation space~\citep{zhao2026sgwam}. 
CogWAM follows this trend toward control-relevant future representations, while conditioning future prediction on an explicit execution-aware Semantic State.

\textbf{Task Progress across Semantic and Control Timescales.}
A related challenge is the mismatch between task-level semantics and the finer temporal scale of physical control. Hierarchical VLAs address this issue through intermediate semantic commands~\citep{shi2025hirobot,intelligence2025pi_05}, while completion-aware approaches track task progress across action chunks~\citep{yang2025seqvla}. WALL-WM further organizes world-action learning around semantically coherent events to address the granularity mismatch among language, visual dynamics, and fixed-length actions~\citep{li2026wallwm}. CogWAM instead decouples Semantic State transitions from fixed-horizon action generation: the state persists across multiple replanning cycles and is updated only when observed progress indicates a semantic transition. This formulation is related to temporal abstraction in reinforcement learning~\citep{sutton1999between}, but operates on a language-level task state whose updates are event-triggered by execution progress~\citep{tabuada2007event}.

\textbf{Agentic Orchestration and System-1/System-2 Robot Policies.}
A growing line of work moves beyond improving a single visuomotor policy and instead couples fast learned control with slower foundation-model reasoning, forming an emerging System-1/System-2 paradigm for robot decision making. HarnessWAM places a VLM-based task manager above a WAM to maintain global task state and trigger deliberation when needed~\citep{gu2026harnesswam}; VLAs-as-Tools lets a high-level VLM invoke specialized VLA policies as executable tools with progress-aware replanning~\citep{lei2026towards}; and recent studies of hierarchical VLA agents systematically examine how planner--controller interfaces, memory, and switching mechanisms affect long-horizon execution~\citep{hu2026matters}. More general agentic systems further extend this idea to policy routing, recovery, and autonomous policy improvement~\citep{huang2026roboharness,li2026roboclaw}. 
At the other extreme, frontier multimodal models have been directly deployed as closed-loop robot policies: GPT-6 Astra is evaluated across all 42 RoboDojo tasks~\citep{zhang2026unexpected}, while concurrent GPT-as-Policy work explores a hybrid setting in which Astra selectively reviews and corrects trajectories generated by a pretrained $\pi_{0.5}$ controller~\citep{su2026astra}. 
Collectively, these systems reflect a broader shift from monolithic robot policies toward architectures that separate high-level semantic reasoning from high-frequency sensorimotor control. CogWAM follows the same separation of semantic and physical timescales, but takes a different route: rather than keeping System-2 reasoning as an external agent that repeatedly plans, routes, or corrects a downstream policy, CogWAM compresses its output into a persistent Semantic State and makes that state part of the learned world-action interface, allowing semantic cognition to persist across multiple System-1 control cycles without being recomputed at every step.

\section{Method}
\label{sec:method}

\subsection{Problem Formulation}
\label{sec:problem_formulation}

We consider language-conditioned robot manipulation from multi-view observations. At timestep $t$, let $\mathcal{O}_t=\{\boldsymbol{I}_t^{(v)}\}_{v=1}^{V}$ denote the current observation, $\ell$ the task instruction, and $\boldsymbol{a}_t$ an action chunk of horizon $H$. A standard visuomotor policy models $p_\theta(\boldsymbol{a}_t\mid\mathcal{O}_t,\ell)$. 

For long-horizon manipulation, however, the current observation may not reveal which task events have already been completed or which subtask is currently active.
CogWAM therefore maintains a persistent Semantic State $S_{t^-}=(m_{t^-},s_{t^-})$, where $m_{t^-}$ records completed task events and $s_{t^-}$ denotes the active subtask. Given $\mathcal{O}_t$, $\ell$, and $S_{t^-}$, the model first determines the Semantic State $S_t$ consistent with the current task progress.

\begin{figure*}[t]
    \centering
    \includegraphics[
        width=\textwidth,
        pagebox=cropbox,
        clip
    ]{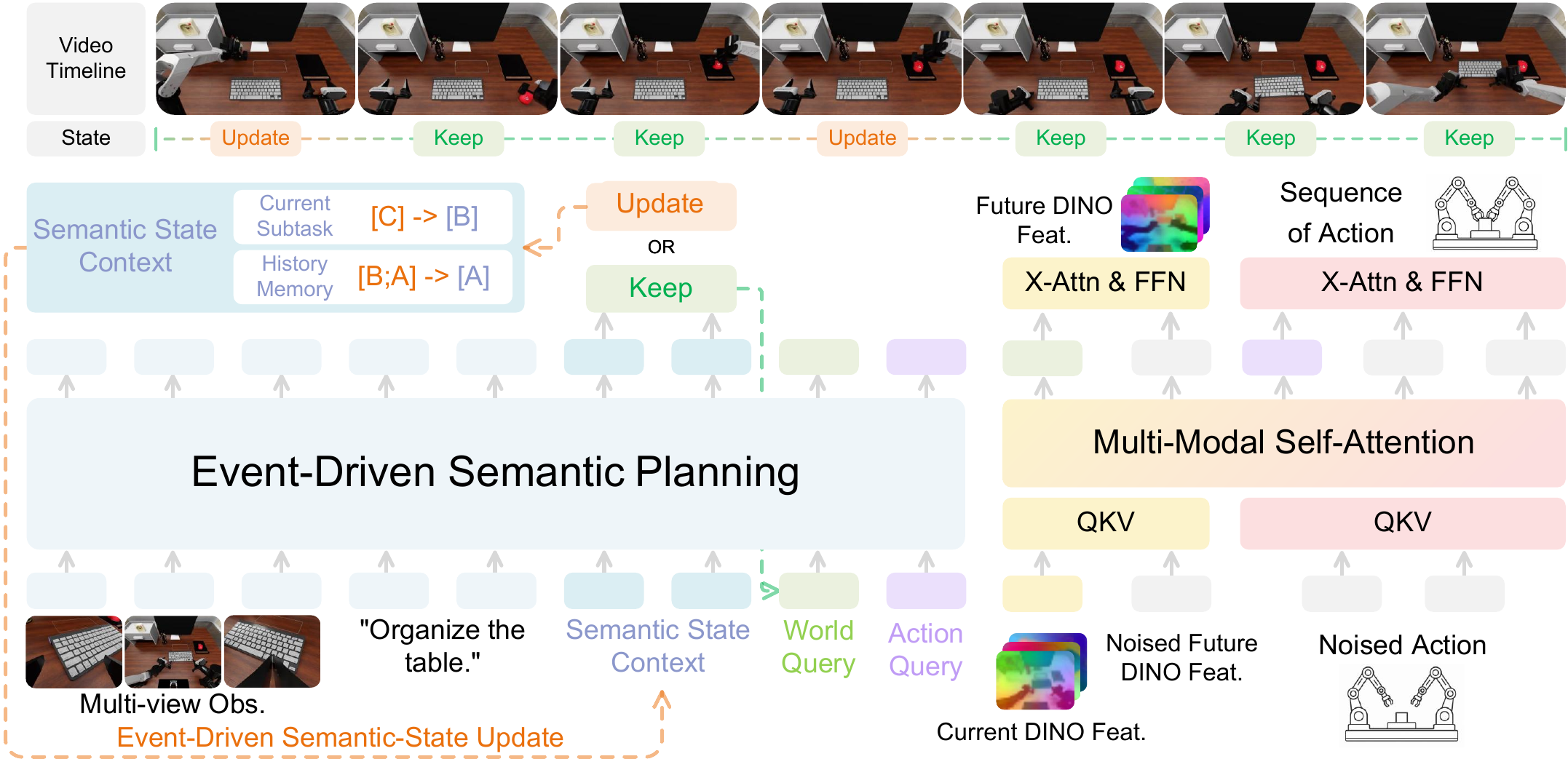}
    \caption{\textbf{Architecture of CogWAM.} 
    An event-triggered \texttt{KEEP}/\texttt{UPDATE} mechanism maintains a persistent Semantic State during closed-loop execution. WORLD and ACTION queries condition parallel branches for future multi-view DINO feature prediction and continuous action generation.}    
    \label{fig:cogwam_overview}
\end{figure*}

During training, CogWAM additionally predicts a future multi-view latent representation $\mathcal{Z}_{t+H}=\{\boldsymbol{z}_{t+H}^{(v)}\}_{v=1}^{V}$. We factorize the conditional distribution as
\begin{equation}
\label{eq:cogwam_factorization}
p_\theta\!\left(
S_t,\mathcal{Z}_{t+H},\boldsymbol{a}_t
\mid \mathcal{O}_t,\ell,S_{t^-}
\right) =
p_\theta\!\left(
S_t\mid\mathcal{O}_t,\ell,S_{t^-}
\right)
p_\theta\!\left(
\mathcal{Z}_{t+H},\boldsymbol{a}_t
\mid\mathcal{O}_t,\ell,S_t
\right).
\end{equation}
The first factor models task-progress transitions, while the second models progress-conditioned world prediction and action generation. Importantly, action generation does not depend on the predicted future representation; future-world prediction is used only as a training-time objective.
At inference time, the world branch is removed, and CogWAM directly predicts $p_\theta(\boldsymbol{a}_t\mid\mathcal{O}_t,\ell,S_t)$ from the current observation and maintained Semantic State. The overall architecture is shown in Fig.~\ref{fig:cogwam_overview}.

\subsection{Event-Triggered Semantic State}
\label{sec:semantic_planning}
CogWAM maintains a persistent Semantic State throughout task execution and updates it only when visual evidence indicates a change in task progress. As a language-level summary of execution progress, the Semantic State preserves task history that may no longer be recoverable from the current observation. Instead of regenerating semantic context at every step~\cite{yang2026world}, CogWAM separates transition detection from state generation.

\textbf{Semantic-State Transition.}
Let $d_t\in\{\texttt{KEEP},\texttt{UPDATE}\}$ denote whether the stored Semantic State remains valid at timestep $t$. Given the current observation $\mathcal{O}_t$, task instruction $\ell$, and previous state $S_{t^-}$, the vision-language model (VLM)~\cite{li2026rynnbrain} predicts:
\begin{equation}
d_t
\sim
p_\theta\!\left(
d_t
\mid
\mathcal{O}_t,\ell,S_{t^-}
\right).
\label{eq:semantic_decision}
\end{equation}
When $d_t=\texttt{UPDATE}$, the model generates a memory increment $\Delta m_t$ describing the newly completed task event and the updated active subtask $s_t$:
\begin{equation}
(\Delta m_t,s_t)
\sim
p_\theta\!\left(
\Delta m_t,s_t
\mid
\mathcal{O}_t,\ell,S_{t^-},
d_t=\texttt{UPDATE}
\right).
\label{eq:semantic_generation}
\end{equation}
The resulting Semantic State is updated as:
\begin{equation}
S_t=
\begin{cases}
S_{t^-},
& d_t=\texttt{KEEP},\\
(m_{t^-}\oplus\Delta m_t,s_t),
& d_t=\texttt{UPDATE},
\end{cases}
\label{eq:semantic_transition}
\end{equation}
where $\oplus$ appends the newly completed event to the stored memory. Thus, $d_t$ determines when the Semantic State changes, while $(\Delta m_t,s_t)$ determines how it changes. The state can persist across multiple action chunks until observed progress supports a transition, decoupling semantic evolution from fixed control horizons.
At inference, $d_t$ is predicted in a single forward pass. A \texttt{KEEP} decision avoids autoregressive semantic generation, whereas an \texttt{UPDATE} decision additionally generates $(\Delta m_t,s_t)$. Details of semantic-state supervision are provided in Appendix~\ref{app:semantic_supervision}.

\subsection{Progress-Conditioned World--Action Interface}
\label{sec:semantic_queries}
Maintaining task progress alone does not determine how semantic context should influence different physical prediction objectives. Future-world prediction and action generation capture complementary aspects of robot behavior: the former models scene evolution, while the latter generates executable control trajectories. CogWAM therefore introduces progress-conditioned WORLD and ACTION queries to transfer the same task-progress context to both branches while preserving objective-specific representations.

Given the maintained Semantic State $S_t$, CogWAM appends two sets of learnable query tokens after the current observation, task instruction, and semantic context in the causal sequence. The resulting hidden representations $\boldsymbol{H}_t^w$ and $\boldsymbol{H}_t^a$ are generated by the WORLD and ACTION queries:
\begin{equation}
\boldsymbol{H}_t^w,
\boldsymbol{H}_t^a
=
f_\theta(
\mathcal{O}_t,\ell,S_t,
Q^{w},Q^{a}
).
\end{equation}
The causal ordering allows both query sets to attend to the current observation, instruction, and Semantic State, while their outputs are routed to different downstream objectives. Specifically, $\boldsymbol{H}_t^w$ conditions future multi-view latent prediction, whereas $\boldsymbol{H}_t^a$ conditions continuous action generation. Their branch-specific supervision encourages the WORLD and ACTION queries to extract task-relevant information for their respective prediction targets, enabling shared task-level semantics with specialized physical representations.
This design is supported by the gradient analysis in Appendix~\ref{app:query_analysis}, which shows that World and Action objectives share perceptual grounding but require increasingly distinct representations at the query readout stage.

\subsection{Latent World--Action Modeling}
\label{sec:world_action}
Given the progress-conditioned representations $\boldsymbol{H}_t^w$ and $\boldsymbol{H}_t^a$, CogWAM jointly learns future-world prediction and continuous action generation over the action horizon. 

\textbf{Multi-View Future Latents.}
A frozen DINO encoder $E_{\mathrm{DINO}}$ maps each current image $\boldsymbol{I}_t^{(v)}$ and its future counterpart $\boldsymbol{I}_{t+H}^{(v)}$ to latent features $\boldsymbol{z}_t^{(v)}$ and $\boldsymbol{z}_{t+H}^{(v)}$. Across $V$ views, we denote the current and future representations as $\mathcal{Z}_t=\{\boldsymbol{z}_t^{(v)}\}_{v=1}^{V}$ and $\mathcal{Z}_{t+H}=\{\boldsymbol{z}_{t+H}^{(v)}\}_{v=1}^{V}$.
Predicting future DINO features avoids appearance-level reconstruction while preserving object structure and spatial information relevant to physical state changes~\cite{nilaksh2026reconstruction,nie2026larylatentactionrepresentation}. Multi-view targets further capture complementary global and local interaction cues~\cite{torne2026mem}.

\textbf{World and Action Branches.}
CogWAM instantiates the world and action branches as two streams in a Mixture-of-Transformers (MoT) architecture~\citep{liang2024mixture}. Both streams share $\mathcal{Z}_t$ as the current visual anchor and interact through multimodal self-attention, while retaining modality-specific cross-attention and feed-forward blocks. The World stream predicts $\mathcal{Z}_{t+H}$ conditioned on $\boldsymbol{H}_t^w$, whereas the Action stream generates $\boldsymbol{a}_t$ conditioned on $\boldsymbol{H}_t^a$.
During joint training, a structured attention mask prevents Action tokens from attending to future-latent tokens, avoiding target leakage and ensuring that action generation uses only information available at inference~\cite{yuan2026fast}. Future-latent prediction therefore serves as a training-time co-objective rather than an intermediate plan for action generation. At inference, the World stream is removed, leaving a direct action path conditioned on $\mathcal{Z}_t$ and $\boldsymbol{H}_t^a$.

\subsection{Training Strategy}
\label{sec:training}
\textbf{Joint Training Objective.}
CogWAM jointly optimizes action generation, future-latent prediction, and Semantic State modeling. For the action branch, Gaussian noise $\boldsymbol{\epsilon}^{a}\sim\mathcal{N}(\boldsymbol{0},\boldsymbol{I})$ is interpolated with the target action chunk $\boldsymbol{a}_t$ at flow time $\tau\sim\mathcal{U}(0,1)$. The world branch analogously interpolates $\boldsymbol{\epsilon}^{w}\sim\mathcal{N}(\boldsymbol{0},\boldsymbol{I})$ with the future representation $\mathcal{Z}_{t+H}$:
\begin{equation}
\boldsymbol{a}_{t,\tau}
=
(1-\tau)\boldsymbol{a}_t+\tau\boldsymbol{\epsilon}^{a},
\qquad
\mathcal{Z}_{t+H,\tau}
=
(1-\tau)\mathcal{Z}_{t+H}+\tau\boldsymbol{\epsilon}^{w}.
\label{eq:flow_interpolation}
\end{equation}
Conditioned on the current visual representation and the corresponding progress-conditioned queries, the two branches predict their velocity fields:
\begin{align}
\mathcal{L}_{\mathrm{action}}
&=
\mathbb{E}
\left[
\left\|
\boldsymbol{v}_{\theta}^{a}
\left(
\boldsymbol{a}_{t,\tau},\tau;
\mathcal{Z}_t,\boldsymbol{H}_t^a
\right)
-
\left(
\boldsymbol{\epsilon}^{a}-\boldsymbol{a}_t
\right)
\right\|_2^2
\right],\\
\mathcal{L}_{\mathrm{world}}
&=
\mathbb{E}
\left[
\left\|
\boldsymbol{v}_{\theta}^{w}
\left(
\mathcal{Z}_{t+H,\tau},\tau;
\mathcal{Z}_t,\boldsymbol{H}_t^w
\right)
-
\left(
\boldsymbol{\epsilon}^{w}-\mathcal{Z}_{t+H}
\right)
\right\|_2^2
\right].
\label{eq:world_action_flow_matching}
\end{align}
Semantic State prediction uses a decision loss $\mathcal{L}_{\mathrm{dec}}$ for \texttt{KEEP}/\texttt{UPDATE} classification and an autoregressive loss $\mathcal{L}_{\mathrm{gen}}$ for memory increment and updated subtask. Given annotated decision $d_t^\star$,
\begin{equation}
\mathcal{L}_{\mathrm{sem}}
=
\mathcal{L}_{\mathrm{dec}}
+
\mathbf{1}\!\left[
d_t^\star=\texttt{UPDATE}
\right]
\mathcal{L}_{\mathrm{gen}}.
\label{eq:semantic_objective}
\end{equation}
The complete training objective is
\begin{equation}
\mathcal{L}
=
\mathcal{L}_{\mathrm{action}}
+
\lambda_{\mathrm{world}}\mathcal{L}_{\mathrm{world}}
+
\lambda_{\mathrm{sem}}\mathcal{L}_{\mathrm{sem}},
\label{eq:total_objective}
\end{equation}
where $\lambda_{\mathrm{world}}$ and $\lambda_{\mathrm{sem}}$ control the contributions of future-latent and semantic supervision.

\textbf{Boundary-Aware Semantic Sampling.}
Semantic transitions are sparse, so uniform sampling strongly favors \texttt{KEEP}. We therefore oversample \texttt{UPDATE} events together with hard-\texttt{KEEP} examples near annotated transition boundaries. These neighboring observations are visually similar to transition states but do not yet justify advancing the subtask, encouraging the model to localize semantic transitions more precisely. This rebalancing is applied only to semantic supervision; world and action training follow the original demonstration distribution.

\textbf{Progressive Semantic Conditioning.}
Training the world and action branches only with annotated Semantic States creates a train--inference mismatch, since deployment conditions on model-predicted states. We therefore use the predicted state $\widehat{S}_t$ as semantic context with probability $\rho_k$ and the annotated state $S_t^\star$ otherwise, where $\rho_k$ increases with training progress $k$. Semantic prediction remains supervised by the annotated target; only the context used to construct $\boldsymbol{H}_t^w$ and $\boldsymbol{H}_t^a$ is progressively replaced. This exposes both branches to the semantic context encountered during closed-loop inference.

\section{Experiments and Analysis}
\label{sec:experiments}

We evaluate CogWAM in both simulation and the real world. Following a question-driven evaluation protocol, our experiments are designed to answer the following four questions:
\begin{enumerate}[label=\textbf{Q\arabic*:}, leftmargin=*, topsep=2pt, itemsep=1pt, parsep=0pt, partopsep=0pt]
    \item How does CogWAM perform on multi-task robotic manipulation benchmarks?
    \item Does the Semantic State improve long-horizon manipulation?
    \item How should the Semantic State be updated during closed-loop execution?
    \item How well does CogWAM transfer to real-world robotic manipulation?
\end{enumerate}

\subsection{Experimental Setup}
\label{sec:experimental_setup}

\textbf{Benchmarks.}
We conduct simulation experiments on RoboDojo~\cite{chen2026robodojo} and BiCoord~\cite{peng2026bicoord}, both under the multi-task setting. RoboDojo evaluates manipulation capabilities across Generalization, Precision, Long-Horizon, Memory, and Open categories. BiCoord contains 18 long-horizon bimanual tasks with stage-wise annotations and an average of 4.27 stages per task, providing a complementary test bed for multi-stage bimanual coordination.

\textbf{Metrics.}
For RoboDojo, we report Score and Success Rate (SR) for each capability dimension and average them across the five dimensions. Following the official protocol, each task is evaluated over 50 episodes; Generalization uses 25 standard and 25 randomized episodes. For BiCoord, we report task Success Rate (SR) and Stage-wise Success Rate (SSR) over 100 rollouts per task. For the update-strategy analysis, we additionally report semantic-model calls, inference latency, and transition delay relative to annotated stage boundaries, measured in the real-robot deployment where the control frequency imposes a fixed replanning budget.

\subsection{Simulation Experiments}
\label{sec:simulation_experiments}
\textbf{Q1: Multi-task benchmark performance.}
We compare CogWAM with representative VLA and WAM baselines on RoboDojo and BiCoord under the multi-task setting.

\definecolor{lightgray}{HTML}{EDF8FE}

\begin{table*}[t]
    \centering
    \caption{
    \textbf{Multi-Task Evaluation on RoboDojo.}
    Each cell reports Score / Success Rate (in \%).
    Methods are grouped by whether prior embodied robot-data pre-training is used before RoboDojo-specific training.
    \textbf{Bold} denotes the best result within each group.
    }
    \label{tab:robodojo_main}

    \resizebox{\textwidth}{!}{
    \setlength{\tabcolsep}{0.58em}
    \begin{tabular}{l|cccccc}
        \toprule
        Method &
        Generalization &
        Precision &
        Long-Horizon &
        Memory &
        Open &
        Average \\
        \midrule

        \multicolumn{7}{l}{\textit{w/ Prior Embodied Robot-Data Pre-training}} \\
        \midrule

        X-VLA~\cite{zheng2026x}
        & 10.47 / 6.78
        & 18.32 / 12.00
        & 16.53 / 9.75
        & 4.76 / 3.56
        & 0.55 / 0.50
        & 10.13 / 6.52 \\

        InternVLA-A1.5~\cite{ma2026internvla}
        & 10.35 / 6.83
        & 15.23 / 10.17
        & 23.80 / 13.75
        & 4.93 / 3.56
        & 1.43 / 1.42
        & 11.15 / 7.14 \\

        $\pi_{0.5}$~\cite{intelligence2025pi_05}
        & 13.38 / 8.17
        & 12.40 / 5.50
        & 23.54 / 14.67
        & 5.89 / 4.67
        & 1.98 / 1.67
        & 11.44 / 6.93 \\

        Spatial Forcing~\cite{li2026spatial}
        & 14.12 / 9.34
        & 17.32 / 10.58
        & 23.26 / 14.58
        & 5.43 / 4.11
        & 1.78 / 1.58
        & 12.38 / 8.04 \\

        Hy-Embodied-0.5-VLA~\cite{zhang2026hy}
        & 11.78 / 8.39
        & 13.81 / 8.00
        & 25.74 / 14.92
        & 13.37 / 12.11
        & 0.65 / 0.58
        & 13.07 / 8.80 \\

        Xiaomi-Robotics-1~\cite{team2026xiaomi}
        & \textbf{23.54 / 17.00}
        & 26.69 / 18.83
        & 38.39 / 23.67
        & 7.81 / 6.56
        & \textbf{3.94 / 3.58}
        & 20.07 / 13.93 \\

        GalaxeaVLA (G0.5)~\cite{liu2026g0}
        & 18.46 / 12.83
        & \textbf{28.25 / 20.42}
        & \textbf{44.12 / 32.25}
        & 8.61 / 7.33
        & 1.73 / 1.58
        & 20.23 / 14.88 \\

        DM0.5~\cite{dexmal2026dm05}
        & 15.77 / 10.95
        & 24.82 / 16.75
        & 33.70 / 19.50
        & \textbf{47.74 / 47.44}
        & 2.43 / 2.08
        & \textbf{24.90 / 19.34} \\

        \midrule
        \multicolumn{7}{l}{\textit{w/o Prior Embodied Robot-Data Pre-training}} \\
        \midrule

        Fast-WAM~\cite{yuan2026fast}
        & 2.33 / 1.11
        & 1.96 / 0.00
        & 9.14 / 5.17
        & 3.55 / 3.44
        & 0.42 / 0.42
        & 3.48 / 2.03 \\

        AHA-WAM~\cite{cai2026aha}
        & 5.79 / 3.27
        & 5.86 / 2.42
        & 8.61 / 2.67
        & 2.97 / 2.78
        & 0.88 / 0.83
        & 4.82 / 2.39 \\

        StarVLA-$\alpha$~\cite{ye2026starvla}
        & 3.94 / 2.33
        & 9.90 / 4.33
        & 14.15 / 6.50
        & 3.34 / 2.44
        & 0.67 / 0.58
        & 6.40 / 3.24 \\

        X-WAM~\cite{guo2026xwam}
        & 7.39 / 3.33
        & 6.72 / 1.83
        & 17.47 / 9.08
        & 6.32 / 4.67
        & 0.57 / 0.25
        & 7.69 / 3.83 \\

        Fast-WAM + CogWAM Interface
        & 12.28 / 9.00
        & 19.61 / 13.50
        & 23.69 / 14.75
        & \textbf{9.17 / 8.00}
        & 1.93 / 1.75
        & 13.33 / 9.40 \\

        \rowcolor{lightgray}
        \textbf{CogWAM (Ours)}
        & \textbf{15.53 / 12.17}
        & \textbf{24.45 / 19.00}
        & \textbf{28.14 / 19.00}
        & 7.65 / 6.33
        & \textbf{2.05 / 2.00}
        & \textbf{15.56 / 11.70} \\

        \bottomrule
    \end{tabular}
    }
\end{table*}
%

\definecolor{lightgreen}{HTML}{EDF8FE}

\begin{table*}[t]
    \centering
    \caption{
    \textbf{Multi-Task Evaluation on BiCoord.}
    Each cell reports Stage-wise Success Rate / Success Rate (in \%). We show the six tasks with the longest average expert trajectories, while \textbf{Avg.} is computed over all 18 tasks. Complete per-task results are provided in Appendix Tab.~\ref{tab:bicoord_main}. 
    }
    \label{tab:bicoord_LH}

    \resizebox{\textwidth}{!}{
        \setlength{\tabcolsep}{0.55em}
        \begin{tabular}{l|ccccccc|c}
        \toprule
        Method
        & \begin{tabular}[c]{@{}c@{}}Clean\\Table\end{tabular}
        & Cook
        & \begin{tabular}[c]{@{}c@{}}Exchange\\Mics\end{tabular}
        & \begin{tabular}[c]{@{}c@{}}Exchange\\Pots\end{tabular}
        & \begin{tabular}[c]{@{}c@{}}Match Blocks\\With Signs\end{tabular}
        & \begin{tabular}[c]{@{}c@{}}Put Objects\\Cabinet\end{tabular}
        & $\cdots$
        & \begin{tabular}[c]{@{}c@{}}Average\\(18 Tasks)\end{tabular} \\
        \midrule

        RDT~\cite{liu2025rdt}
        & 17.5 / 0.0
        & 31.0 / 9.0
        & 35.0 / 23.0
        & \textbf{96.0 / 92.0}
        & 7.0 / 1.0
        & 49.0 / 31.0
        & $\cdots$
        & 34.8 / 16.9 \\

        OpenVLA-OFT~\cite{kim2025fine}
        & 25.0 / 2.0
        & 25.0 / 10.0
        & \textbf{67.5 / 66.0}
        & 56.0 / 53.0
        & 8.7 / 5.0
        & 49.5 / 26.0
        & $\cdots$
        & 36.5 / 23.1 \\

        $\pi_0$~\cite{black2024pi_0}
        & 46.2 / 6.0
        & 29.5 / 14.0
        & 59.5 / 52.0
        & 60.5 / 52.0
        & \textbf{18.7 / 8.0}
        & 24.0 / 6.0
        & $\cdots$
        & 43.2 / 27.2 \\

        $\pi_{0.5}$~\cite{intelligence2025pi_05}
        & 61.5 / 24.0
        & 48.0 / 31.0
        & 62.0 / 55.0
        & 68.0 / 61.0
        & 15.0 / 6.0
        & 58.5 / 42.0
        & $\cdots$
        & 50.8 / 35.6 \\

        \midrule

        \rowcolor{lightgreen}
        \textbf{CogWAM (Ours)}
        & \textbf{80.0 / 50.0}
        & \textbf{77.5 / 52.0}
        & 56.5 / 49.0
        & 76.5 / 70.0
        & 17.3 / 7.0
        & \textbf{84.0 / 74.0}
        & $\cdots$
        & \textbf{58.0 / 43.8} \\

        \bottomrule
        \end{tabular}
    }
\end{table*}

\textbf{RoboDojo.}
As shown in Tab.~\ref{tab:robodojo_main}, CogWAM achieves an average Score/SR of \textbf{15.56/11.70\%} without prior embodied robot-data pre-training, substantially outperforming existing methods under the same setting. Relative to StarVLA-$\alpha$~\cite{ye2026starvla}, it improves the average Score by \textbf{9.16} and SR by \textbf{8.46} percentage points, with particularly strong gains in Precision and Long-Horizon manipulation. CogWAM also surpasses several methods that rely on prior embodied robot-data pre-training.

\textbf{BiCoord.}
Tab.~\ref{tab:bicoord_LH} further evaluates long-horizon bimanual manipulation. Averaged over all 18 tasks, CogWAM achieves an SSR/SR of \textbf{58.0/43.8\%}, outperforming the strongest evaluated baseline, $\pi_{0.5}$~\cite{intelligence2025pi_05}, by \textbf{7.2/8.2} percentage points. Its strongest gains occur on sustained multi-stage tasks such as Clean Table, Cook, and Put Objects Cabinet, while performance is less uniformly improved on tasks dominated by low-level bimanual coordination. Together, the two benchmarks show that CogWAM substantially strengthens multi-task and long-horizon execution without relying on prior robot-data pre-training.

\textbf{Interface and Representation Analysis.}
We further analyze two design choices underlying CogWAM: whether the proposed semantic interface transfers across world-action architectures, and how the predictive representation affects performance.
(1) Cross-architecture transfer.
Applying the same semantic interface to Fast-WAM~\cite{yuan2026fast} improves its average Score/SR from
\textbf{3.48/2.03\%} to \textbf{13.33/9.40\%}, while retaining the original Fast-WAM world-action formulation.
This substantial gain shows that the benefit of the semantic interface is not specific to CogWAM's underlying WAM architecture.
(2) Predictive representation and model efficiency. We compare predictive formulations while keeping the semantic interface unchanged. Fast-WAM repurposes a pretrained Wan2.2-TI2V-5B video backbone~\cite{wan2025}, whereas CogWAM predicts future DINOv3 features with a 0.44B world stream trained from scratch. Despite the reduced world-modeling capacity, CogWAM improves average Score/SR from \textbf{13.33/9.40\%} to \textbf{15.56/11.70\%}, with gains on Generalization, Precision, and Long-Horizon, while Wan2.2 remains stronger on Memory. This suggests that control-relevant visual representations can provide effective predictive supervision without large generative backbones. Since representation, capacity, and initialization differ jointly, we interpret this comparison as evidence of formulation efficiency rather than isolated representation quality.

\begin{figure*}[t]
    \centering
    \includegraphics[width=\linewidth]{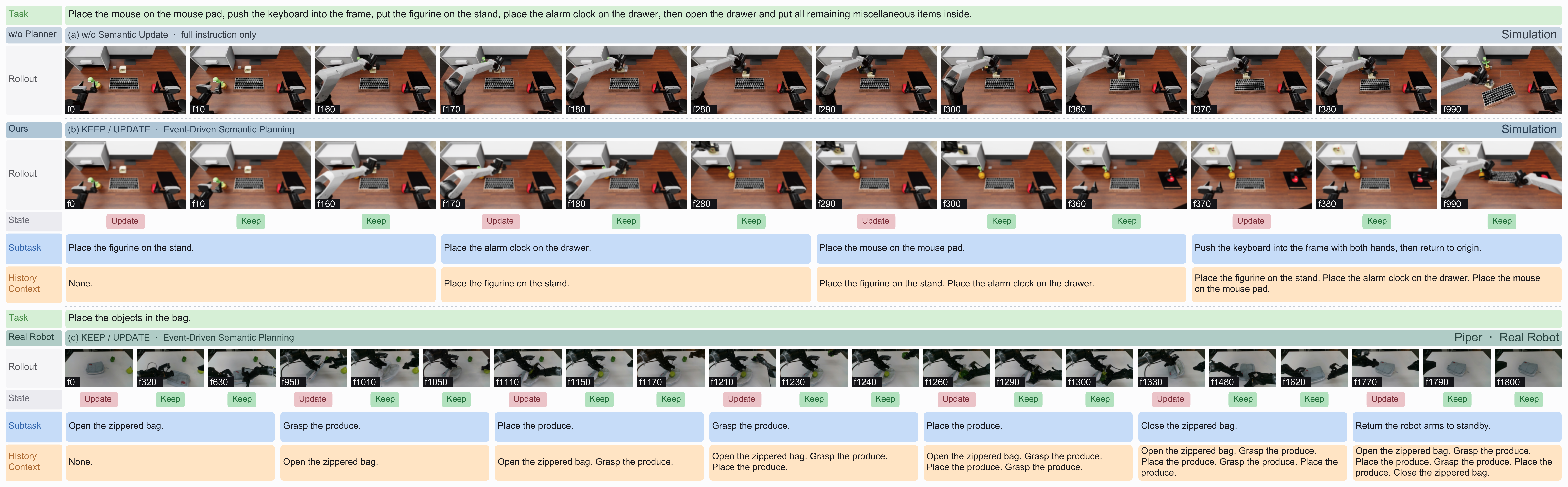}
    \caption{\textbf{Qualitative Analysis of Semantic State Progression.}
    Simulation rollouts illustrate how persistent Semantic States track task progress across action chunks, reducing subtask drift, repetition, and premature transitions.
    The real-world rollout visualizes the predicted \texttt{KEEP}/\texttt{UPDATE} decisions together with the corresponding Semantic States throughout execution.}
    \label{fig:subtask_pred}
\end{figure*}

\begin{figure*}[t]
    \centering
    \includegraphics[width=\linewidth]{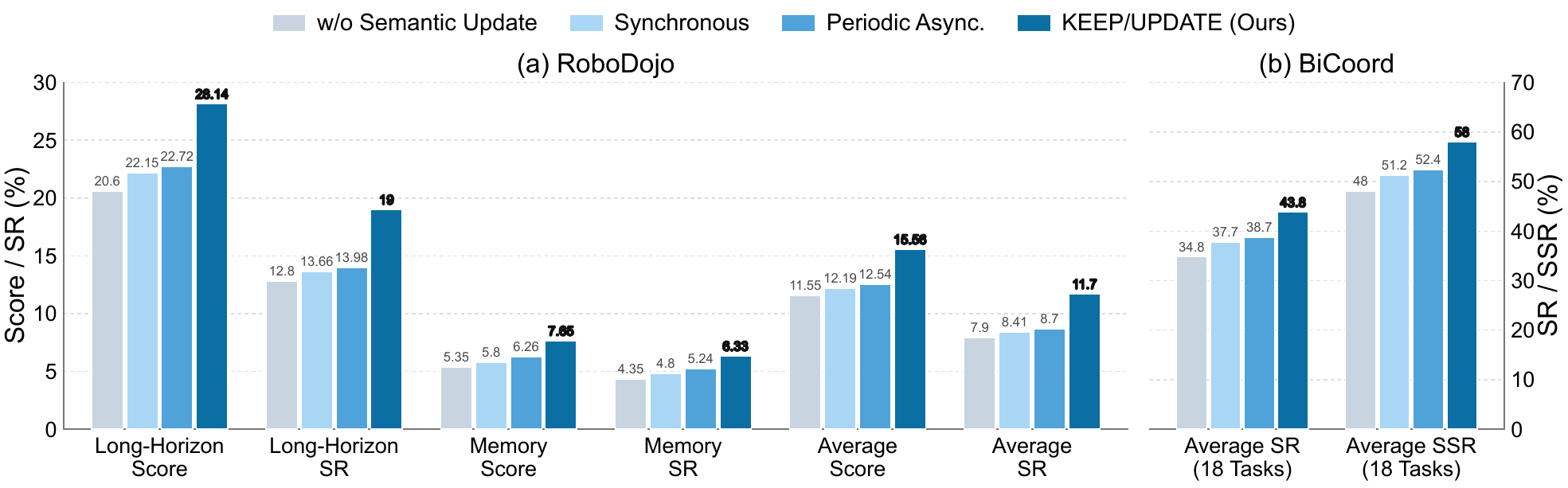}
    \caption{\textbf{Effect of Semantic State and Update Strategies.}
    Task performance on (a) RoboDojo (Score / SR) and (b) BiCoord (SR / SSR), shown with benchmark-specific vertical scales.
    \textit{w/o Semantic State} conditions the policy only on the current observation and global task instruction, while the remaining variants differ in how the Semantic State is updated.}
    \label{fig:semantic_update}
\end{figure*}

\textbf{Q2: Effect of the Semantic State.}
We isolate the contribution of the Semantic State by comparing CogWAM with a world-action MoT counterpart that shares the same backbone and training setting but conditions on only the current observation and global task instruction, without any persistent Semantic State. As illustrated in Fig.~\ref{fig:subtask_pred}, this baseline is more prone to subtask drift, repetition of completed stages, and premature transitions. In contrast, CogWAM preserves the active subtask across multiple action chunks and advances it only when execution progress changes. Fig.~\ref{fig:semantic_update} further quantifies these gains on the Long-Horizon and Memory categories of RoboDojo and the stage-wise progression metrics of BiCoord.

\textbf{Q3: Semantic State update strategies.}
We compare three update strategies: \textit{Synchronous}, \textit{Asynchronous}, and our \textit{Event-triggered} update. Synchronous updating regenerates the Semantic State at every replanning step and therefore incurs substantial computation. Asynchronous updating reduces this overhead by running the semantic planner at a fixed lower frequency, but its updates can be misaligned with actual subtask transitions. CogWAM instead performs a lightweight \texttt{KEEP}/\texttt{UPDATE} decision and regenerates the Semantic State only when a semantic transition is detected. As shown in Fig.~\ref{fig:semantic_update}, the event-triggered strategy performs best on both benchmarks; notably, synchronous updating does not surpass asynchronous updating despite regenerating far more often, indicating that update timing matters more than update frequency. We further compare semantic-model calls, inference latency, and transition delay on the real robot (Sec.~\ref{sec:real_world}).

\subsection{Real-World Experiments}
\label{sec:real_world}

\setlength{\abovecaptionskip}{1pt}

\begin{wraptable}{r}{0.55\textwidth}
\small

\caption{\textbf{Real-world evaluation.} Success counts across basic manipulation tasks and distribution shifts.}
\label{tab:real_world}

\setlength{\tabcolsep}{3.5pt}
\begin{tabular}{lc|lc}
\toprule
\textbf{Basic Task} & \textbf{Success} &
\textbf{Generalization} & \textbf{Success} \\
\midrule
Place Objects      & 20 / 20 & Spatial Location  & 21 / 30 \\
Organize Utensils  & 17 / 20 & Object Appearance & 20 / 30 \\
Put in Drawer      & 18 / 20 & Distractor         & 17 / 30 \\
Fill Pen Holder    & 18 / 20 & Novel Objects      & 14 / 30 \\
Place in Bag       & 16 / 20 & \multicolumn{2}{c}{--} \\
Block Sorting      &  5 / 20 & \multicolumn{2}{c}{--} \\
\midrule
\rowcolor{lightgray}
\textbf{Total} & \textbf{94 / 120} &
\textbf{Total} & \textbf{72 / 120} \\
\bottomrule
\end{tabular}
\end{wraptable}

\begin{figure*}[t]
\centering
\includegraphics[width=\textwidth]{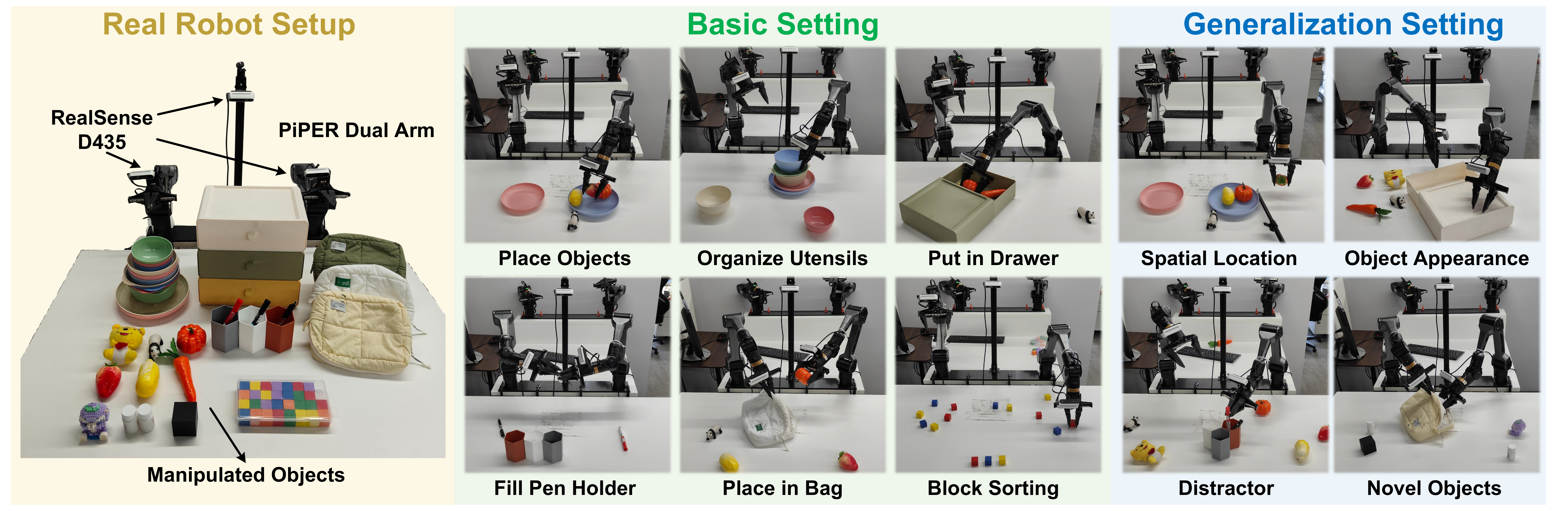}
\caption{\textbf{Real-world setup and task examples.} CogWAM is deployed on a dual-arm AgileX PIPER platform with two 6-DoF manipulators and parallel grippers, using synchronized RGB observations from one fixed head-mounted and two wrist-mounted Intel RealSense D435 cameras.}
\label{fig:real_world_setup}
\end{figure*}

\textbf{Q4: Real-world evaluation.}
We evaluate CogWAM in the real world along two complementary tracks: task performance on six manipulation tasks and robustness under four controlled distribution shifts. Fig.~\ref{fig:real_world_setup} illustrates the robotic setup and representative task examples. Details of data collection, optimization, compute, and deployment hyperparameters are provided in Appendix~\ref{app:training}.

\textbf{Evaluation Protocol.}
We evaluate CogWAM on six multi-stage tabletop tasks: \textit{Place Objects}, \textit{Organize Utensils}, \textit{Put in Drawer}, \textit{Fill Pen Holder}, \textit{Place in Bag}, and \textit{Block Sorting}, covering repeated manipulation, category-level organization, articulated and deformable object interaction, bimanual coordination, and precision placement. For the basic setting, each task is evaluated over 20 closed-loop trials with predefined initial configurations, yielding 120 trials in total. We further assess robustness under four controlled distribution shifts, each evaluated independently across all six tasks. For each shift dimension, we perform 5 closed-loop trials per task, yielding 30 trials per dimension. Tab.~\ref{tab:real_world} summarizes the successful trials across both settings. We additionally visualize a representative \textit{Place in Bag} rollout, showing the predicted \texttt{KEEP}/\texttt{UPDATE} decisions and corresponding subtask memory in Fig.~\ref{fig:subtask_pred}.

\begin{table}[t]
\centering
\small
\caption{
\textbf{Deployment cost of update strategies.}
\textit{Total} denotes semantic-model calls per rollout, while \textit{Regen.} counts calls that regenerate the Semantic State; \textit{Dec.\ tok.} denotes tokens decoded per replanning step. Transition delay is the lag between a subtask boundary and the first replanning step at which the strategy can observe it. Latency is measured on an NVIDIA RTX 5090.
}
\label{tab:update_strategy}

\setlength{\tabcolsep}{4pt}
\begin{tabular}{lccccc}
\toprule
& \multicolumn{2}{c}{\textbf{Semantic-model calls}} & \multicolumn{2}{c}{\textbf{Inference latency}} & \textbf{Transition} \\
\cmidrule(lr){2-3}\cmidrule(lr){4-5}
\textbf{Update strategy} & Total & Regen.\ $\downarrow$ & Mean (ms) $\downarrow$ & Dec.\ tok.\ $\downarrow$ & \textbf{delay (ms)} $\downarrow$ \\
\midrule
Synchronous            & 65.4 & 65.4                    & 927.8          & 45          & \textbf{152} \\
Asynchronous (1\,Hz)   & 22.1 & 22.1                    & 309.3          & 45          & 502 \\
\midrule
\rowcolor{lightgray}
\textbf{Event-triggered (ours)} & 65.4 & \textbf{\phantom{0}4.0} & \textbf{146.7} & \textbf{\phantom{0}0} & \textbf{152} \\
\bottomrule
\end{tabular}
\end{table}

\textbf{Real-time efficiency of semantic updating.}
The real robot runs a 30\,Hz control loop and issues one replanning request every 10 control steps, leaving a 333\,ms budget per replanning step. On \textit{Place in Bag}, five held-out rollouts average 21.6\,s, 65.4 replanning steps, and 4.00 semantic transitions per episode. As shown in Tab.~\ref{tab:update_strategy}, synchronous updating achieves a 152\,ms transition delay but incurs 927.8\,ms latency, exceeding the budget by $2.8\times$. Asynchronous updating reduces latency to 309.3\,ms by updating every 30 control steps, but increases transition delay to 502\,ms. In contrast, CogWAM evaluates the \texttt{KEEP}/\texttt{UPDATE} decision at every replanning step, as frequently as synchronous updating, but requires $16.4\times$ fewer Semantic State regenerations. The \texttt{KEEP}/\texttt{UPDATE} decision is predicted from the semantic module's final hidden state without autoregressive decoding, enabling frequent transition checks at low cost. This yields the same 152\,ms transition delay as synchronous updating, while reducing inference latency from 927.8\,ms to 146.7\,ms.

\section{Model Analysis}

\subsection{Analysis of VLM-World-Action Interface}
\label{app:wam_analysis}

CogWAM bridges high-level semantic reasoning and low-level physical control through a structured World--Action interface, adding a predictive World branch alongside the Action branch that conventional policies optimize alone~\citep{intelligence2025pi_05}. This section asks why the two branches need separate queries (\S\ref{app:query_analysis}), what future-world prediction contributes beyond direct action supervision (\S\ref{app:world_ablation}), and whether the interface transfers to a different World--Action backbone (\S\ref{app:interface_transfer}).

\subsubsection{Semantic Interface between VLM and World-Action Learning}
\label{app:query_analysis}

Future-world prediction and action generation share the same instruction and observation, but supervise the model through different output modalities: one learns a representation of scene evolution, the other a representation of executable control. Two design questions follow. \textbf{How deep into the shared VLM should the objectives remain coupled, and through which interface should each read out the representation it needs?} We answer both by measuring, rather than assuming, how the two objectives interact inside the trunk.

Treating the gradient direction as the update an objective requests from a parameter block, we report
\begin{equation}
\mathrm{GA}(\theta)
=
\frac{
\nabla_{\theta}\mathcal{L}_{\mathrm{world}}
\cdot
\nabla_{\theta}\mathcal{L}_{\mathrm{action}}
}{
\left\|\nabla_{\theta}\mathcal{L}_{\mathrm{world}}\right\|_2
\left\|\nabla_{\theta}\mathcal{L}_{\mathrm{action}}\right\|_2
},
\qquad
\mathrm{MR}(\theta)
=
\frac{\left\|\nabla_{\theta}\mathcal{L}_{\mathrm{world}}\right\|_2}
     {\left\|\nabla_{\theta}\mathcal{L}_{\mathrm{action}}\right\|_2},
\label{eq:ga}
\end{equation}
over depth-ordered blocks $\theta$ of the shared VLM trunk. $\mathrm{GA}$ asks whether the objectives request the same update direction, $\mathrm{MR}$ whether the world objective carries enough magnitude to matter. We compute both measures from the unweighted world and action losses in Eq.~\ref{eq:ga}. At a fixed checkpoint, multiplying either gradient by a positive scalar does not change $\mathrm{GA}$. Changing the loss weights during training may, however, change the learned parameters and gradient directions, so this scale invariance alone does not make $\mathrm{GA}$ directly comparable across differently trained checkpoints. To attribute any difference to the interface alone, we compare the two architectures on a matched configuration that varies only in how the world and action branches attach to the VLM: both use a RynnBrain1.1 backbone~\cite{li2026rynnbrain}, action horizon 16, the same RoboDojo recipe and the same frozen DINOv3-B teacher. This controlled pair is therefore not the final model of Tab.~\ref{tab:app_arch}, which uses the RynnBrain1.1 backbone at horizon 25; we trade the exact deployment configuration for a clean attribution. All measurements use 120 training batches at 20k steps, report medians with 95\% bootstrap intervals, and split the 24-layer language-model stack into equal thirds (layers $0$--$7$, $8$--$15$, $16$--$23$).

\begin{figure}[t]
\centering
\includegraphics[width=0.9\linewidth]{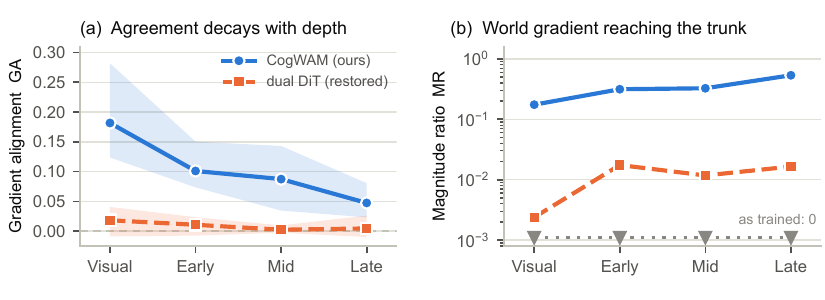}
\caption{
\textbf{World and Action gradients inside the shared VLM, by depth.}
(a) Gradient alignment, with $95\%$ bootstrap intervals over $120$ batches. CogWAM is significantly positive at every depth and decays inward from the visual tower; the dual-DiT baseline with its gradient path restored is indistinguishable from zero throughout.
(b) World-gradient magnitude relative to the action gradient. As trained, the dual-DiT baseline detaches the world branch, so no world gradient reaches the trunk and $\mathrm{GA}$ is undefined.
}
\label{fig:gradient_alignment}
\end{figure}

\textbf{Shared grounding, specialized abstraction.} Sec.~\ref{sec:semantic_queries} motivates separate WORLD and ACTION queries as a way to share task-level semantics while preserving objective-specific representations; the gradients make that trade-off measurable. In CogWAM the objectives are positively but weakly aligned throughout the trunk, and the alignment decreases with depth (Fig.~\ref{fig:gradient_alignment}a): every interval excludes zero, and the medians fall monotonically from $0.182\,[0.124,0.282]$ in the visual tower through $0.101$ and $0.087$ to $0.047\,[0.023,0.080]$ in the last language-model third ($0.081$ over the trunk as a whole). Because the blocks are measured on the same batches, a paired test applies; the visual tower exceeds every language-model block (Wilcoxon $p<10^{-3}$), and LM-early exceeds LM-late; the median of the batch-wise paired differences is $0.023\,[0.011,0.046]$, although adjacent blocks are not always separable. This is an expectation over batches rather than a per-batch property: 66\% to 76\% of batches are positive, so there is no systematic opposition between the objectives, but individual batches do disagree. Both objectives must recover the same objects, geometry and scene state, which makes the perceptual trunk genuinely shared; by the semantic depth at which the queries read out, the representation each requires is largely its own. Notably the deeper blocks receive proportionally more world gradient (Fig.~\ref{fig:gradient_alignment}b, $\mathrm{MR}$ rising from $0.17$ to $0.53$) while agreeing with the action objective less, so the divergence is one of direction rather than of the world signal fading out. A single query would have to summarise two only marginally related requirements within one fixed bottleneck, whereas separate WORLD and ACTION queries let each objective retain the full subspace for roughly $3\times10^{4}$ additional parameters.

\textbf{External heads in the evaluated dual-DiT configuration.} The dual-DiT baseline attaches independent world and action heads to a standard VLM. As trained, its recipe detaches the world branch from the backbone, so $\|\nabla_{\theta}\mathcal{L}_{\mathrm{world}}\|$ is identically zero on every VLM block and $\mathrm{GA}$ is undefined: the world modality shapes only its own head. Since that detachment is a choice in the recipe rather than a consequence of using external heads, we also measure the counterfactual in which the path is restored. With the gradient path restored, the evaluated dual-DiT baseline has a whole-trunk $\mathrm{GA}$ of $0.012\,[-0.005,0.022]$, with every per-block interval covering zero. Its $\mathrm{MR}$ is $0.02$, compared with $0.37$ for CogWAM, indicating a substantially smaller world-gradient contribution to the shared trunk. In this evaluated configuration, the external world head contributes little to shared representation learning even when its gradient path is restored.

\textbf{Query ordering.} Sec.~\ref{sec:world_action} relies on a structured attention mask to keep ACTION tokens from attending to future-latent tokens; Fig.~\ref{fig:attention_mask} shows the forward information path. The zero world-loss gradient on ACTION query parameters is consistent with this query ordering, while the attention mask directly prevents action tokens from accessing future-latent inputs. The WORLD queries receive gradients from both objectives ($\mathrm{MR}=2.63$), with positive alignment between them ($\mathrm{GA}=0.152$).

\begin{tcolorbox}[title=Key Findings: Designing a VLM--World-Action Interface]
An effective interface should therefore not hand one unified semantic representation to every downstream objective, but preserve shared context at the depth where the objectives agree and let each specialise where that agreement runs out.
\end{tcolorbox}

\subsubsection{Decomposing CogWAM}
\label{app:world_ablation}

We next decompose CogWAM into the two additions that distinguish it from action-only control: predictive world supervision and semantic conditioning. Tab.~\ref{tab:wo_world} follows the progression
\textit{Action MoT} $\rightarrow$ \textit{MoT} $\rightarrow$ \textit{CogWAM}.
Action MoT retains the World--Action architecture but is trained only with the action objective. MoT additionally learns to predict future observations in the frozen DINO feature space. CogWAM further introduces the semantic interface, which conditions the policy on VLM-derived task progress and maintains an event-triggered Semantic State containing the active subtask and accumulated task history.

\begin{table*}[h]
\centering
\small
\setlength{\tabcolsep}{1.8pt}
\renewcommand{\arraystretch}{1.08}

\caption{
\textbf{Stepwise decomposition of CogWAM on RoboDojo.}
Each cell reports Score / Success Rate (in \%).
Action MoT $\rightarrow$ MoT adds DINO-space future-world prediction;
MoT $\rightarrow$ CogWAM further adds the CogWAM Interface.
}
\label{tab:wo_world}

\begin{tabular*}{\textwidth}{
    @{\extracolsep{\fill}}
    l|ccc|cccccc
    @{}
}
    \toprule
    &
    \multicolumn{3}{c|}{\textbf{Components}} &
    \multicolumn{6}{c}{\textbf{RoboDojo Evaluation}} \\
    \cmidrule(lr){2-4}
    \cmidrule(lr){5-10}

    Method &
    Pred. &
    VLM &
    State &
    Gen. &
    Prec. &
    Long-H. &
    Mem. &
    Open &
    Avg. \\
    \midrule

    Action MoT
    & --
    & --
    & --
    & 8.00 / 5.20
    & 18.20 / 12.60
    & 21.62 / 12.00
    & 3.50 / 2.00
    & \textbf{2.10} / 1.90
    & 10.68 / 6.74 \\

    MoT
    & $\checkmark$
    & --
    & --
    & 11.73 / 8.83
    & 21.24 / 14.00
    & 21.88 / 13.64
    & 5.72 / 4.67
    & 1.05 / 1.00
    & 12.32 / 8.43 \\

    \rowcolor{lightgray}
    \textbf{CogWAM}
    & $\checkmark$
    & $\checkmark$
    & $\checkmark$
    & \textbf{15.53 / 12.17}
    & \textbf{24.45 / 19.00}
    & \textbf{28.14 / 19.00}
    & \textbf{7.65 / 6.33}
    & 2.05 / \textbf{2.00}
    & \textbf{15.56 / 11.70} \\

    \bottomrule
\end{tabular*}

\vspace{0.3ex}
\footnotesize
\textit{Pred.}: DINO-space future-world prediction;
\textit{VLM}: VLM-derived task-progress conditioning;
\textit{State}: event-triggered Semantic State.
\end{table*}

\textbf{Future prediction primarily improves perceptual grounding.}
Moving from Action MoT to MoT raises Generalization by $+3.73$ and Precision by $+3.04$ Score, while Long-Horizon changes by only $+0.26$. This pattern is consistent with future-world prediction acting mainly as a representation-learning signal: predicting how the scene will evolve encourages the shared representation to preserve objects, geometry, and interaction-relevant visual structure that are useful for selecting the next action. It does not, however, explicitly encode where the robot is within a multi-stage task, and therefore provides little direct mechanism for resolving progress ambiguity over long horizons.

\textbf{The semantic interface reduces action uncertainty by conditioning on task progress.}
Moving from MoT to CogWAM produces the largest gain on Long-Horizon ($+6.26$ Score), but also improves Generalization ($+3.80$), Precision ($+3.21$), Memory ($+1.93$), and Open ($+1.00$). These broader gains are expected because semantic conditioning is useful beyond tasks that are nominally long-horizon. From a conditional-prediction perspective, an observation and instruction alone may admit several plausible action modes when the same visual configuration can occur at different stages of a task. Introducing the active subtask and task history supplies an additional condition $z_t$, replacing the harder prediction problem
$p(a_t \mid o_t,\ell)$
\(\)
with
$p(a_t \mid o_t,\ell,z_t)$.
Equivalently, the interface is intended to reduce the conditional uncertainty of the action distribution,
$H(A_t\mid O_t,\ell,z_t) \leq H(A_t\mid O_t,\ell)$,
whenever task progress disambiguates otherwise plausible behaviors.

This reduction is most consequential in long-horizon tasks, where perceptually similar states can recur after different histories and therefore require different subsequent actions. The accumulated Semantic State explicitly records which task events have already occurred, reducing this form of history-dependent state aliasing. The same conditioning can nevertheless benefit shorter-horizon Generalization and Precision tasks: specifying the current semantic stage narrows the set of admissible behaviors and gives the action model a more stable task-relevant context, rather than requiring it to infer both task phase and motor command from the observation at every replanning step.

Importantly, the MoT-to-CogWAM comparison adds the complete semantic interface---VLM-derived task-progress conditioning together with the event-triggered Semantic State. We therefore attribute the improvement in Tab.~\ref{tab:wo_world} to the semantic interface as a whole, rather than claiming that this ablation isolates either component individually. The disproportionately large Long-Horizon gain is consistent with the additional value of explicit task history, but separating current-subtask conditioning from accumulated semantic memory would require a further component-level ablation.

Unlike generative world models that optimize future reconstruction, CogWAM predicts future representations in a frozen DINO feature space~\citep{simeoni2025dinov3}. Future prediction is used only as a training signal, so the World branch need not synthesize pixels at inference time. The compact target instead encourages the shared representation to capture task-relevant scene evolution while keeping the predictive objective lightweight.

\subsubsection{Transferability of the CogWAM Interface}
\label{app:interface_transfer}

The decomposition above evaluates the CogWAM Interface within our DINO-based World--Action formulation. We next ask whether its benefit depends on this particular predictive backbone. To test this, we attach the same interface to Fast-WAM~\citep{yuan2026fast}, while retaining Fast-WAM's original world-action formulation, Wan2.2-TI2V-5B predictive backbone~\citep{wan2025}, and action decoder.

\begin{table*}[t]
\centering
\small
\setlength{\tabcolsep}{2.5pt}
\renewcommand{\arraystretch}{1.08}

\caption{
\textbf{Transferability of the CogWAM Interface on RoboDojo.}
Each cell reports Score / Success Rate (in \%).
Applying the same semantic interface to Fast-WAM substantially improves
performance without changing its underlying predictive formulation.
}
\label{tab:interface_transfer}

\resizebox{\textwidth}{!}{
\begin{tabular}{l|cccccc}
\toprule
\textbf{Method} &
\textbf{Gen.} &
\textbf{Prec.} &
\textbf{Long-H.} &
\textbf{Mem.} &
\textbf{Open} &
\textbf{Avg.} \\
\midrule

Fast-WAM~\cite{yuan2026fast}
& 2.33 / 1.11
& 1.96 / 0.00
& 9.14 / 5.17
& 3.55 / 3.44
& 0.42 / 0.42
& 3.48 / 2.03 \\

Fast-WAM + CogWAM Interface
& 12.28 / 9.00
& 19.61 / 13.50
& 23.69 / 14.75
& \textbf{9.17 / 8.00}
& 1.93 / 1.75
& 13.33 / 9.40 \\

\rowcolor{lightgray}
\textbf{CogWAM}
& \textbf{15.53 / 12.17}
& \textbf{24.45 / 19.00}
& \textbf{28.14 / 19.00}
& 7.65 / 6.33
& \textbf{2.05 / 2.00}
& \textbf{15.56 / 11.70} \\

\bottomrule
\end{tabular}
}
\end{table*}

Adding the CogWAM Interface raises Fast-WAM's average Score/SR from $3.48/2.03$ to $13.33/9.40$. The improvement is broad rather than confined to one capability: Generalization increases by $+9.95$ Score, Precision by $+17.65$, Long-Horizon by $+14.55$, Memory by $+5.62$, and Open by $+1.51$. This transfer is important because the underlying predictive model is unchanged; the gain therefore does not rely on CogWAM's DINO-space future prediction.

Together with Tab.~\ref{tab:wo_world}, the result separates two roles. Predictive world modeling shapes the representation of how the scene may evolve, while the CogWAM Interface conditions action generation on task progress and execution history. In conditional-prediction terms, the former enriches the predictive representation of the observation, whereas the latter supplies an additional task-state variable that narrows the set of plausible continuations. This explains why the interface improves not only Long-Horizon tasks, where history-dependent ambiguity is strongest, but also Generalization and Precision tasks, where explicit task-phase conditioning can reduce ambiguity in the next-action distribution.

The remaining gap between Fast-WAM + CogWAM Interface and CogWAM should not be attributed to any single predictive component, since the two systems retain different world-action formulations. The transfer experiment supports the narrower conclusion that the CogWAM Interface is portable: its benefit persists when attached to a substantially different predictive backbone.

\subsection{Language Grounding under Visual Perturbations}
\label{sec:language_grounding}

While the previous analysis studies how semantic information is transferred from the VLM to the World-Action model, we further investigate \textbf{whether the CogWAM Interface enables fine-grained language grounding between instructions and visual entities}.

\begin{figure}[t]
    \centering
    \includegraphics[width=\linewidth]{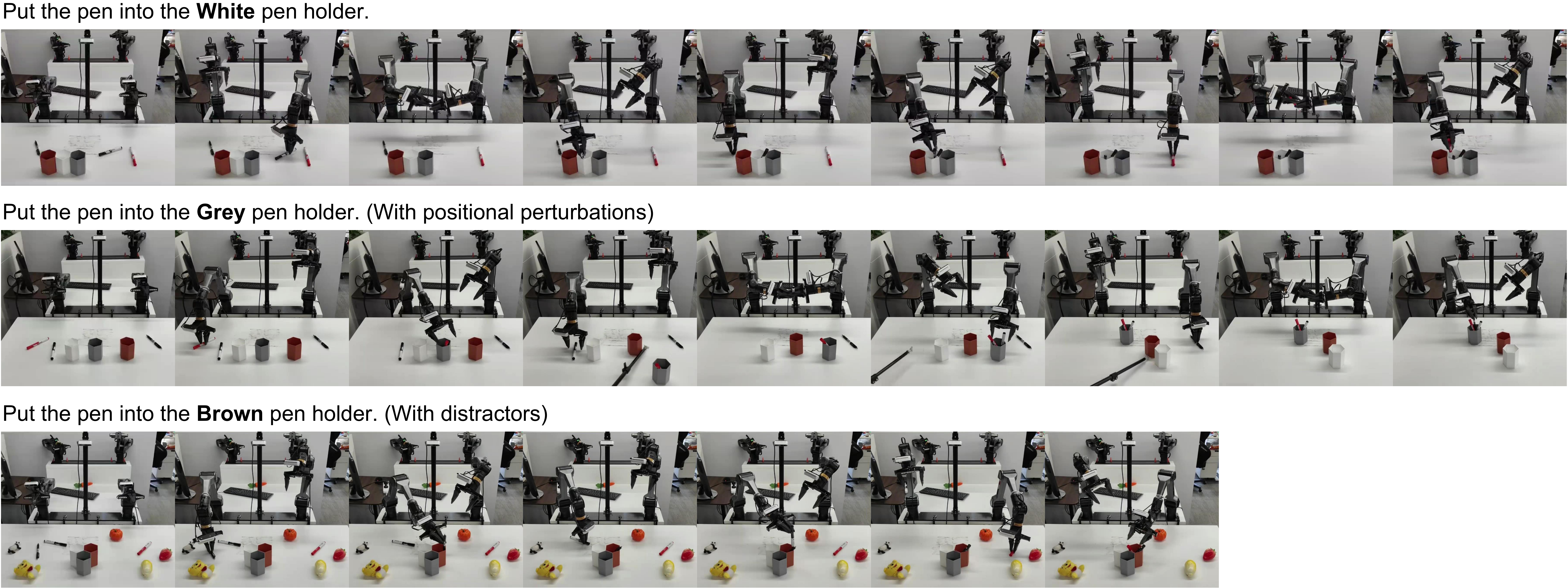}
    \caption{
    \textbf{Language grounding under visual perturbations.}
    CogWAM follows instructions specifying different target pen holders: white, grey, and brown. The grey-holder setting introduces spatial perturbations of the target objects, while the brown-holder setting introduces additional distractor objects. Despite these variations, CogWAM consistently grounds the instructed target and completes the corresponding manipulation behavior.
    }
    \label{fig:instruction_following}
\end{figure}

Fig.~\ref{fig:instruction_following} qualitatively evaluates language grounding on the \textit{Fill Pen Holder} task. We vary the target specified by the instruction while keeping the underlying manipulation objective unchanged. The target object is changed among different pen holders with distinct visual appearances, including white, grey, and brown holders. Furthermore, we introduce two types of visual perturbations: spatial rearrangement of the target objects and additional distractor objects in the scene.

CogWAM successfully identifies the instructed target across these variations, demonstrating robust grounding between linguistic descriptions and visual observations. This behavior highlights a capability provided by the VLM Interface beyond task-progress modeling: instead of relying solely on a compact
language embedding associated with demonstrations, the policy receives a semantically grounded representation that explicitly links instructions with the current visual context.

\begin{tcolorbox}[title=Key Findings: VLM Enables Robust Language Grounding]
Beyond task-progress modeling, the VLM interface grounds language instructions to visual entities, allowing CogWAM to robustly follow fine-grained references under visual variations.
\end{tcolorbox}

\section{Conclusion and Discussion}
\label{sec:Conclusion}


CogWAM shows that explicitly aligning semantic task progress with world-action learning improves execution and deployment efficiency. Across simulation and real-world dual-arm manipulation, it consistently improves task completion without prior embodied robot-data pre-training. Importantly, transferring the same semantic interface to Fast-WAM also yields substantial gains, suggesting that the interface is not tied to a particular WAM architecture and can serve as a reusable bridge between VLM-style semantic reasoning and WAM-style predictive learning.
Our analysis further shows that both update timing and predictive representation matter. Event-triggered updating remains aligned with task transitions while requiring far fewer Semantic State regenerations than synchronous updating. Meanwhile, CogWAM's compact DINO-based predictive formulation provides additional gains over the Fast-WAM-based variant, indicating that semantic alignment and predictive representation are complementary. 
A remaining limitation is that the Semantic State is append-only and cannot revise committed semantic errors. Future work could explore uncertainty-aware state revision and broader transfer across heterogeneous VLM and WAM backbones.

\newpage


\clearpage

\bibliography{paper}
\bibliographystyle{unsrtnat}


\clearpage
\appendix

\renewcommand{\thefigure}{A\arabic{figure}}
\renewcommand{\thetable}{A\arabic{table}}

\setcounter{figure}{0}
\setcounter{table}{0}

\appendix

\section{Implementation Details}
\label{app:implementation}

\subsection{Model Architecture}
\label{app:architecture}

CogWAM couples a vision-language backbone~\citep{li2026rynnbrain}, a Mixture-of-Transformers (MoT) architecture~\citep{liang2024mixture} whose world and action streams are conditioned separately, and a frozen visual teacher~\citep{simeoni2025dinov3}. Tab.~\ref{tab:app_arch} summarizes the three components.

\begin{table}[h]
\centering
\small
\caption{
\textbf{Model architecture.}
The backbone row reports its 24-layer language model; its vision tower is also 24 layers.
The two MoT streams traverse the same 30 layers but keep separate widths, attention projections, and feed-forward blocks; attention head dimension is $128$ for both.
At inference, the world-prediction stream and future-target encoding are removed, while the frozen DINO encoder is retained to extract current visual features.
}
\label{tab:app_arch}
\setlength{\tabcolsep}{6pt}
\begin{tabular}{lcccc}
\toprule
& \textbf{VLM backbone} & \multicolumn{2}{c}{\textbf{World--Action MoT}} & \textbf{Visual teacher} \\
\cmidrule(lr){3-4}
& RynnBrain1.1-2B & World stream & Action stream & DINOv3 ViT-B/16 \\
\midrule
Parameters        & 2.72\,B & 442\,M & 1.01\,B & 85.7\,M \\
Trained           & yes     & yes    & yes     & frozen  \\
Layers            & 24      & \multicolumn{2}{c}{30}     & 12 \\
Attention heads   & \phantom{0}8 & \multicolumn{2}{c}{24} & 12 \\
Hidden dim        & 2048    & \phantom{0}512  & 1024 & \phantom{0}768 \\
Feed-forward dim  & 6144    & 2048   & 4096    & 3072 \\
\midrule
\multicolumn{5}{l}{Learned planner queries: 16 \texttt{WORLD} $+$ 25 \texttt{ACTION} at dim 2048 ($0.084$\,M)} \\
\multicolumn{5}{l}{Trainable total: $4.19$\,B incl.\ projection layers \quad\textbullet\quad Full model incl.\ frozen teacher: $4.27$\,B} \\
\bottomrule
\end{tabular}
\end{table}

\begin{figure*}[h]
\centering
\includegraphics[width=0.9\linewidth]{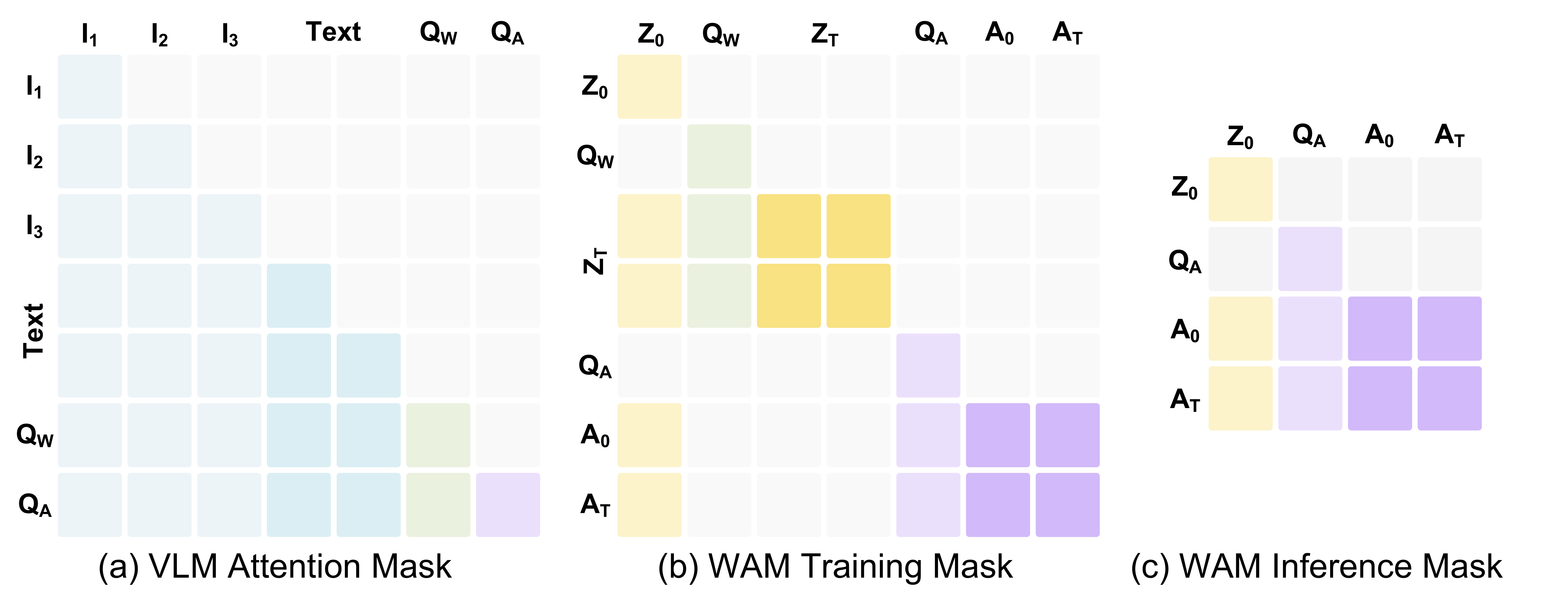}
\caption{
\textbf{Illustration of attention masks in CogWAM.}}
\label{fig:attention_mask}
\end{figure*}

The attention masks used in CogWAM are illustrated in Fig.~\ref{fig:attention_mask}.
For the VLM backbone (Fig.~\ref{fig:attention_mask}(a)), image tokens follow bidirectional attention, while text tokens and learned WORLD/ACTION queries adopt causal attention.
During World--Action MoT training (Fig.~\ref{fig:attention_mask}(b)), future-latent tokens attend to the WORLD conditioning representations and current visual features, while action tokens are masked from future-latent tokens to avoid information leakage.
At inference, the world stream is removed, leaving the action stream conditioned on current visual features and ACTION representations.

\section{Training Details}
\label{app:training}

\subsection{RoboDojo Training Recipe}
\label{app:recipe}

\begin{table}[h]
\centering
\small
\caption{
\textbf{Training hyperparameters.}
The semantic stream is a second, independent dataloader: every step trains both a physical batch and a balanced semantic batch.
}
\label{tab:app_recipe}
\setlength{\tabcolsep}{8pt}
\begin{tabular}{ll}
\toprule
\textbf{Configuration} & \textbf{Value} \\
\midrule
Optimizer                   & AdamW, $\beta = (0.9, 0.95)$, $\epsilon = 10^{-8}$ \\
Weight decay                & $10^{-8}$ \\
Learning rate (VLM backbone) & $1\times10^{-5}$ \\
Learning rate (planner queries) & $1\times10^{-4}$ \\
Learning rate (action model) & $1\times10^{-4}$ \\
LR schedule                 & cosine, $2{,}000$ warmup steps, floor $5\times10^{-7}$ \\
Gradient clipping           & $1.0$ (global norm) \\
Training steps              & $50{,}000$ \\
Batch size (physical)       & $12$ / device, $768$ global \\
Batch size (semantic)       & $6$ / device (2 \texttt{UPDATE} / 2 hard \texttt{KEEP} / 2 random \texttt{KEEP}) \\
Compute                     & $8\times8$ NVIDIA H20 \\
Precision                   & bfloat16, DeepSpeed ZeRO-2, gradient checkpointing \\
Random seed                 & $42$ \\
\midrule
Loss                        & $\mathcal{L}_{\mathrm{action}} + \lambda_{\mathrm{world}}\mathcal{L}_{\mathrm{world}} + \lambda_{\mathrm{sem}}\mathcal{L}_{\mathrm{sem}}$ \\
$\lambda_{\mathrm{world}}$, $\lambda_{\mathrm{sem}}$ & $1.0$, $0.005$ \\
\midrule
Flow-matching schedule      & shifted uniform, $\sigma = su/(1{+}(s{-}1)u)$, $s = 5.0$ \\
Train timesteps             & $1000$ \\
Regression target           & $\epsilon - x_0$ \\
Inference solver            & Euler, $20$ steps, no classifier-free guidance \\
\midrule
Action horizon $H$          & $25$ \\
Future-world stride         & $16$ frames \\
Replanning interval         & $10$ frames \\
Semantic offset             & $-10$ frames \\
Scheduled-sampling $\rho_k$ & $0$ until $30$k, $\to 0.2$ at $40$k, $\to 0.5$ at $50$k \\
Semantic decoding           & greedy, $\leq 96$ new tokens \\
\bottomrule
\end{tabular}
\end{table}

\textbf{Objectives.}
Tab.~\ref{tab:app_recipe} lists the full training recipe used for the RoboDojo benchmark. The pretrained backbone is fine-tuned an order of magnitude more slowly than the randomly initialized queries and action model. The semantic loss is the sum of two next-token cross-entropies over disjoint label spans, one covering the decision token and one the \texttt{Memory Add} and \texttt{Current Subtask} continuation; they are combined without a relative coefficient, since their effective balance comes from the $2$-of-$6$ composition of the semantic batch.

\subsection{Real-World Training Data}
\label{app:data}

\begin{table}[h]
\centering
\small
\caption{
\textbf{Training data.}
Per-task counts after the $5\%$ episode-level validation split.
}
\label{tab:app_data}
\setlength{\tabcolsep}{6pt}
\begin{tabular}{lcccc}
\toprule
\textbf{Task} & \textbf{Train ep.} & \textbf{Train frames} & \textbf{Val ep.} & \textbf{Val frames} \\
\midrule
Place Objects     & \phantom{1{,}0}66 & \phantom{0}83{,}314 & \phantom{0}1 & \phantom{0}1{,}244 \\
Organize Utensils & \phantom{1{,}0}87 & 112{,}181 & \phantom{0}5 & \phantom{0}6{,}725 \\
Put in Drawer     & \phantom{1{,}}437 & 729{,}929 & 23 & 37{,}273 \\
Fill Pen Holder   & 1{,}477 & 220{,}962 & 78 & 10{,}950 \\
Place in Bag      & \phantom{1{,}}954 & 392{,}681 & 49 & 19{,}131 \\
Block Sorting     & 1{,}210 & 410{,}721 & 64 & 21{,}353 \\
\midrule
\textbf{Total}    & \textbf{4{,}231} & \textbf{1{,}949{,}788} & \textbf{220} & \textbf{96{,}676} \\
\bottomrule
\end{tabular}
\end{table}

\textbf{Sampling.}
Episode counts and frame counts rank inversely (Tab.~\ref{tab:app_data}): \textit{Fill Pen Holder} is $34.9\%$ of episodes but $11.3\%$ of frames, \textit{Put in Drawer} $10.3\%$ and $37.4\%$. Sampling over episodes would therefore undersample the long-horizon tasks, so the physical stream samples over frames. Validation is held out at the episode level, so no validation frame is ever seen in training.

\subsection{Real-World Training Progress}
\label{app:progress}

\begin{figure*}[h]
\centering
\includegraphics[width=0.98\linewidth]{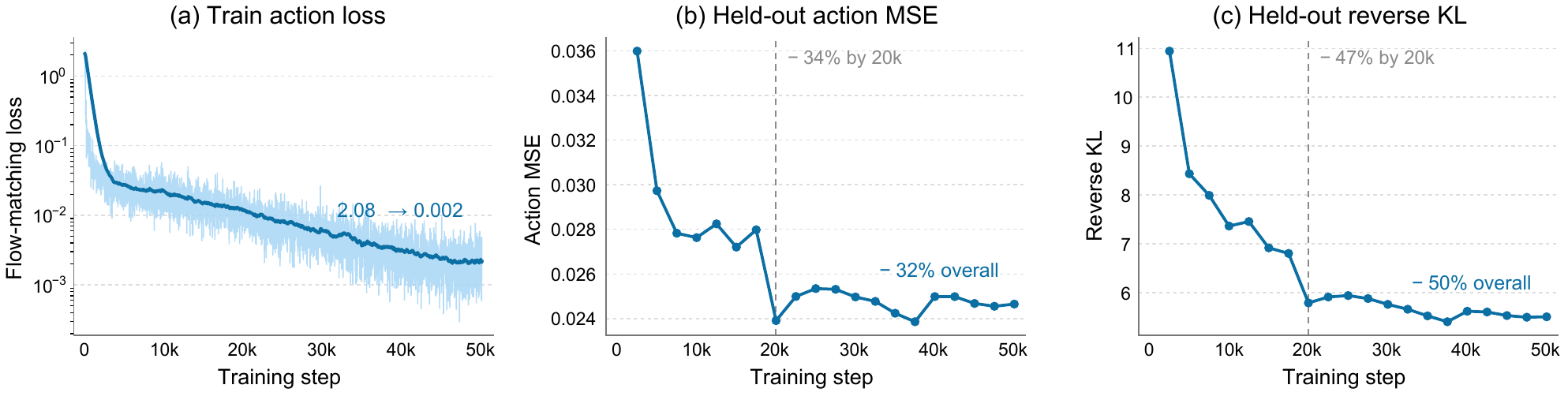}
\caption{
\textbf{Action learning over training.}
Panel (a) shows per-step training loss (light) with an exponential moving average (dark) on a log scale.
Panels (b) and (c) show held-out action MSE and reverse KL at every validation checkpoint.
}
\label{fig:training_progress}
\end{figure*}

Fig.~\ref{fig:training_progress} tracks the action objective over the 50k-step run. Action MSE scores the predicted chunk against the logged one; reverse KL~\citep{generalist2025gen0} asks whether the samples the policy actually draws land near that chunk rather than averaging across modes into an action that matches none of them.

\textbf{Convergence.}
Both held-out measures improve steeply through the first 20k steps and then flatten: action MSE falls $34\%$ by 20k and $32\%$ overall, reverse KL $47\%$ and $50\%$. After 20k the curves separate. Action MSE oscillates within $\pm3\%$ of its final value and reaches its minimum at 37.5k rather than at the end. The remaining fluctuations are small, with no sustained improvement in action MSE. Reverse KL keeps drifting down, indicating that late training sharpens the sampled distribution around the demonstrated action more than it moves the mean prediction. We therefore report 50k as converged rather than as the best of a still-improving run.

\textbf{Relation to closed-loop behaviour.}
Offline action metrics track closed-loop quality across checkpoints, but not across tasks. In informal on-robot checks the 20k and 40k checkpoints behaved comparably in grasp accuracy and generalization while 10k was visibly underfit, consistent with the elbow in Fig.~\ref{fig:training_progress}; we did not run a controlled comparison across checkpoints. Across tasks the picture differs: at 50k, closed-loop success ranges from $20/20$ on \textit{Place Objects} to $5/20$ on \textit{Block Sorting}, a spread the aggregate curves cannot explain, since action MSE compares one predicted chunk to one logged chunk and cannot see whether an error is absorbed at the next replanning step or compounds into a failed grasp. This is also why we track reverse KL: a policy that averages across two viable behaviours can score well on MSE while every sample it draws falls between them, and it is the drawn sample that the robot executes.

\section{Semantic Supervision}
\label{app:semantic_supervision}

This section details the offline procedure that produces the \texttt{KEEP}/\texttt{UPDATE} decision, the memory increment $\Delta m_t$, and the active subtask $s_t$.

\textbf{Offline subtask annotation.}
We generate subtask supervision once offline using GPT-5.6 Sol as a fixed VLM annotator. For each episode, we provide sparsely sampled video frames, the task instruction, and a predefined ordered checklist of semantic stages, and ask the annotator to localize the completion boundary of each stage. The predicted boundaries are mapped back to the original video timeline and converted into right-open subtask spans, yielding the frame-level active subtask $s_t$. For tasks with an episode-dependent number of stages, such as \textit{Arrange Largest Number}, the stage count is first inferred from early observations before constructing the checklist. These annotations are fixed before training; GPT-5.6 Sol is used only for offline annotation and is not involved in policy training or inference.

Some RoboDojo tasks require episode-specific checklist construction. For \textit{Arrange Largest Number}, the number of digit-placement stages varies between four and five; we infer this count from up to three early observations using independent VLM queries followed by majority voting, and instantiate the corresponding checklist before boundary annotation. For tasks with a fixed semantic structure, such as \textit{Spell RoboDojo}, the checklist is fixed in advance. Thus, the VLM annotator localizes transitions between predefined semantic stages rather than freely generating the task decomposition.

\begin{algorithm}[t]
\caption{Offline subtask annotation}
\label{alg:subtask_annotation}
\begin{algorithmic}[1]
\REQUIRE Episode $V=\{x_t\}_{t=1}^{T}$, instruction $\ell$, task $c$
\STATE Retrieve the ordered stage checklist $\mathcal{S}_c$
\IF{$c$ has an episode-dependent stage count}
\STATE Infer the stage count from early observations and instantiate $\mathcal{S}_c$
\ENDIF
\STATE Sample representative observations $\mathcal{O}$ from $V$
\STATE Predict ordered stage boundaries $\mathbf{b}$ from $(\ell,S_c,\mathcal{O})$
\STATE Map $\mathbf{b}$ to the original timeline and construct right-open spans
\RETURN Frame-level active-subtask labels $\{s_t^\star\}_{t=1}^{T}$
\end{algorithmic}
\end{algorithm}

\textbf{Canonicalization.}
The offline annotation procedure above produces right-open segment ends paired with subtask text for each episode. Before constructing semantic supervision, we canonicalize these segments to remove annotation gaps. An unlabelled interval is not retained as a separate segment, because entering and leaving such a gap would otherwise create two artificial semantic transitions. Leading and internal gaps are therefore absorbed into the following labelled segment, while trailing gaps are absorbed into the final labelled segment, ensuring that each semantic stage boundary induces exactly one transition.

\textbf{Decision labels.}
Let $(s_t, m_t)$ denote the active subtask and the cumulative memory of completed subtasks at frame $t$. With a semantic offset matching the replanning interval of $10$ frames,
\begin{equation}
d_t =
\begin{cases}
\texttt{UPDATE}, & \text{if } t < 10, \text{ or } \mathrm{N1}(s_t) \neq \mathrm{N1}(s_{t-10}), \text{ or } \mathrm{N1}(m_t) \neq \mathrm{N1}(m_{t-10}),\\
\texttt{KEEP}, & \text{otherwise,}
\end{cases}
\end{equation}
where $\mathrm{N1}(\cdot)$ collapses whitespace, casefolds, and strips terminal punctuation. Frames with $t < 10$ have no valid cached state and are forced to \texttt{UPDATE}; we exclude them when scoring, since counting them would inflate recall for free.

The memory increment is computed positionally: we take the longest common prefix of the two item lists and emit the remaining suffix of $m_t$. Cumulative memory only grows at the tail, and a set-keyed difference erases a genuine second completion whenever two subtasks share a sentence. On this corpus the set-keyed variant produced an empty increment for $46\%$ and $57\%$ of \texttt{UPDATE} frames on two tasks, which trains the model to emit \texttt{UPDATE} and then write nothing.

\textbf{Boundary-aware sampling.}
Only frames on the replanning phase are eligible. They are partitioned into \texttt{UPDATE} frames, hard \texttt{KEEP} frames one replanning interval before or after an \texttt{UPDATE}, and the remaining random \texttt{KEEP} frames. Hard \texttt{KEEP} frames are visually close to a transition but do not yet support advancing the subtask, so they are where a spurious \texttt{UPDATE} is most likely. Each semantic batch draws $2$ frames from each pool. The training split holds $197{,}617$ eligible frames, $9.19\%$ of them \texttt{UPDATE}; the sampler replays the small pools and subsamples the large one so the model sees a $1{:}1{:}1$ ratio.

\definecolor{lightgreen}{HTML}{EDF8FE}

\begin{table*}[t]
    \centering
    \caption{
    \textbf{Multi-Task Evaluation on BiCoord.}
    Each cell reports Stage-wise Success Rate (SSR) / Success Rate (SR) in \%.
    \textbf{Avg.} is averaged over all 18 tasks.
    \textbf{Bold} denotes the best result in each column.
    }
    \label{tab:bicoord_main}

    \resizebox{\textwidth}{!}{
        \setlength{\tabcolsep}{0.45em}
        \begin{tabular}{l|c|cccccccc}
        \toprule
        Method / Task & Average
        & \begin{tabular}[c]{@{}c@{}}Balance\\Roller\end{tabular}
        & \begin{tabular}[c]{@{}c@{}}Build\\Bridge\end{tabular}
        & \begin{tabular}[c]{@{}c@{}}Build Tower\\With Blocks\end{tabular}
        & \begin{tabular}[c]{@{}c@{}}Clean\\Table\end{tabular}
        & \begin{tabular}[c]{@{}c@{}}Collect\\Pens\end{tabular}
        & \begin{tabular}[c]{@{}c@{}}Cook\end{tabular}
        & \begin{tabular}[c]{@{}c@{}}Divide Block\\Tower\end{tabular}
        & \begin{tabular}[c]{@{}c@{}}Exchange\\Mics\end{tabular} \\
        \midrule

        RDT~\cite{liu2025rdt}
        & 34.8 / 16.9
        & 43.0 / 15.0
        & 50.2 / 47.0
        & 1.0 / 0.0
        & 17.5 / 0.0
        & 68.8 / 15.0
        & 31.0 / 9.0
        & 12.8 / 2.0
        & 35.0 / 23.0 \\

        OpenVLA-OFT~\cite{kim2025fine}
        & 36.5 / 23.1
        & 74.5 / 49.0
        & 2.0 / 2.0
        & 0.0 / 0.0
        & 25.0 / 2.0
        & 46.2 / 5.0
        & 25.0 / 10.0
        & 11.1 / 0.0
        & \textbf{67.5} / \textbf{66.0} \\

        $\pi_0$~\cite{black2024pi_0}
        & 43.2 / 27.2
        & 81.5 / 68.0
        & 47.0 / 44.0
        & 5.0 / 2.0
        & 46.2 / 6.0
        & 81.8 / 47.0
        & 29.5 / 14.0
        & 10.8 / 0.0
        & 59.5 / 52.0 \\

        $\pi_{0.5}$~\cite{intelligence2025pi_05}
        & 50.8 / 35.6
        & 89.8 / 82.0
        & 54.3 / 51.0
        & 21.6 / 11.0
        & 61.5 / 24.0
        & 84.8 / 57.0
        & 48.0 / 31.0
        & 14.8 / \textbf{1.0}
        & 62.0 / 55.0 \\

        \midrule

        \rowcolor{lightgreen}
        \textbf{CogWAM (Ours)}
        & \textbf{58.0} / \textbf{43.8}
        & \textbf{96.5} / \textbf{95.0}
        & \textbf{60.2} / \textbf{58.0}
        & \textbf{35.0} / \textbf{19.0}
        & \textbf{80.0} / \textbf{50.0}
        & \textbf{87.2} / \textbf{66.0}
        & \textbf{77.5} / \textbf{52.0}
        & \textbf{18.0} / \textbf{1.0}
        & 56.5 / 49.0 \\

        \bottomrule
        \end{tabular}
    }

    \vspace{0.6ex}

    \resizebox{\textwidth}{!}{
        \setlength{\tabcolsep}{0.45em}
        \begin{tabular}{cccccccccc}
        \toprule
        \begin{tabular}[c]{@{}c@{}}Exchange\\Pots\end{tabular}
        & \begin{tabular}[c]{@{}c@{}}Extract Bottom\\Block To Top\end{tabular}
        & \begin{tabular}[c]{@{}c@{}}Fetch Block\\With Roller\end{tabular}
        & \begin{tabular}[c]{@{}c@{}}Handover Block\\With Bowls\end{tabular}
        & \begin{tabular}[c]{@{}c@{}}Jigsaw\end{tabular}
        & \begin{tabular}[c]{@{}c@{}}Match Blocks\\With Signs\end{tabular}
        & \begin{tabular}[c]{@{}c@{}}Place Plate\\And Cup\end{tabular}
        & \begin{tabular}[c]{@{}c@{}}Put Objects\\Cabinet\end{tabular}
        & \begin{tabular}[c]{@{}c@{}}Stack\\Bowls\end{tabular}
        & \begin{tabular}[c]{@{}c@{}}Sweep\\Block\end{tabular} \\
        \midrule

        \textbf{96.0} / \textbf{92.0}
        & 25.5 / 17.0
        & 49.5 / 0.0
        & \textbf{20.0} / 0.0
        & 11.5 / 0.0
        & 7.0 / 1.0
        & 70.0 / 40.0
        & 49.0 / 31.0
        & 28.3 / 3.0
        & 10.0 / 10.0 \\

        56.0 / 53.0
        & 55.5 / 53.0
        & 70.5 / 44.0
        & 3.0 / 0.0
        & 25.2 / 0.0
        & 8.7 / 5.0
        & 79.2 / 55.0
        & 49.5 / 26.0
        & 14.0 / 1.0
        & 44.0 / 44.0 \\

        60.5 / 52.0
        & 43.5 / 40.0
        & 69.5 / 43.0
        & 9.0 / \textbf{4.0}
        & 36.5 / 6.0
        & \textbf{18.7} / \textbf{8.0}
        & 66.2 / 25.0
        & 24.0 / 6.0
        & 28.0 / 12.0
        & \textbf{61.0} / \textbf{61.0} \\

        68.0 / 61.0
        & 71.9 / 68.0
        & 71.4 / 45.0
        & 5.7 / 2.0
        & 44.5 / 8.0
        & 15.0 / 6.0
        & 73.8 / 45.0
        & 58.5 / 42.0
        & 34.8 / 18.0
        & 34.0 / 34.0 \\

        \midrule

        \rowcolor{lightgreen}
        76.5 / 70.0
        & \textbf{95.0} / \textbf{95.0}
        & \textbf{73.0} / \textbf{46.0}
        & 3.0 / 0.0
        & \textbf{51.0} / \textbf{9.0}
        & 17.3 / 7.0
        & \textbf{80.2} / \textbf{63.0}
        & \textbf{84.0} / \textbf{74.0}
        & \textbf{40.3} / \textbf{23.0}
        & 12.0 / 12.0 \\

        \bottomrule
        \end{tabular}
    }

\end{table*}


\begin{table*}[h]
    \centering
    \caption{
    \textbf{Complete task-wise results on RoboDojo.}
    Each cell reports Score / Success Rate (in \%).
    For Generalization tasks, \textit{Std.} and \textit{Rand.} denote the
    standard and randomized evaluation settings, respectively.
    Methods are grouped according to whether prior embodied robot-data
    pre-training is used before RoboDojo-specific training.
    Within the group without robot pre-training,
    \textbf{bold} and \underline{underlined} numbers denote the best and
    second-best results for each metric, respectively.
    Methods with robot pre-training are reported for reference and are
    excluded from the ranking.
    Fast-WAM+Interface and \textbf{CogWAM (Ours)} are our own evaluation logs. Bold denotes the best result in each column.
    }
    \label{tab:robodojo_taskwise}

    \scriptsize
    \renewcommand{\arraystretch}{1.0}
    \setlength{\tabcolsep}{0.55em}

    \resizebox{\textwidth}{!}{
    \begin{tabular}{l|ccc>{\columncolor{lightgray}}c|cc}
        \toprule
        &
        \multicolumn{4}{c|}{\textbf{Without Robot Pre-training}} &
        \multicolumn{2}{c}{\textbf{With Robot Pre-training}} \\
        Task &
        Fast-WAM &
        Fast-WAM+Interface &
        StarVLA-$\alpha$ &
        \textbf{CogWAM (Ours)} &
        $\pi_{0.5}$ &
        GalaxeaVLA (G0.5) \\
        \midrule

        \multicolumn{7}{l}{\textbf{Generalization}} \\
        \midrule
        stack\_bowls (Std.) &
        5.87 / 2.67 &
        \textbf{72.20} / \textbf{68.00} &
        \underline{15.27} / \underline{10.67} &
        \textbf{72.20} / \textbf{68.00} &
        76.00 / 72.00 &
        64.07 / 58.67 \\

        stack\_bowls (Rand.) &
        0.80 / 0.00 &
        \underline{3.00} / 0.00 &
        1.20 / 0.00 &
        \textbf{4.80} / 0.00 &
        14.20 / 4.00 &
        20.33 / 13.33 \\

        push\_T (Std.) &
        0.00 / 0.00 &
        0.00 / 0.00 &
        0.00 / 0.00 &
        0.00 / 0.00 &
        0.00 / 0.00 &
        0.00 / 0.00 \\

        push\_T (Rand.) &
        0.00 / 0.00 &
        0.00 / 0.00 &
        0.00 / 0.00 &
        0.00 / 0.00 &
        0.00 / 0.00 &
        0.00 / 0.00 \\

        pack\_objects\_into\_box (Std.) &
        2.87 / 0.00 &
        \underline{14.40} / 0.00 &
        8.07 / \underline{2.67} &
        \textbf{14.60} / \textbf{4.00} &
        22.80 / 4.00 &
        23.80 / 5.33 \\

        pack\_objects\_into\_box (Rand.) &
        0.27 / 0.00 &
        0.40 / 0.00 &
        \underline{1.07} / 0.00 &
        \textbf{7.60} / 0.00 &
        15.40 / 1.33 &
        12.67 / 1.33 \\

        fold\_clothes (Std.) &
        32.00 / 22.67 &
        \underline{53.60} / \underline{52.00} &
        23.47 / 16.00 &
        \textbf{63.20} / \textbf{60.00} &
        48.00 / 38.67 &
        53.87 / 48.00 \\

        fold\_clothes (Rand.) &
        \textbf{2.13} / \textbf{1.33} &
        0.00 / 0.00 &
        \underline{0.53} / 0.00 &
        0.00 / 0.00 &
        16.53 / 9.33 &
        18.67 / 14.67 \\

        hang\_mugs (Std.) &
        2.20 / 0.00 &
        \underline{11.00} / \textbf{4.00} &
        4.33 / 0.00 &
        \textbf{15.20} / \textbf{4.00} &
        9.53 / 0.00 &
        12.47 / 4.00 \\

        hang\_mugs (Rand.) &
        0.00 / 0.00 &
        \textbf{1.20} / 0.00 &
        \underline{0.80} / 0.00 &
        \textbf{1.20} / 0.00 &
        5.00 / 0.00 &
        4.80 / 0.00 \\

        sweep\_blocks (Std.) &
        0.00 / 0.00 &
        \textbf{4.00} / \textbf{4.00} &
        \underline{1.33} / \underline{1.33} &
        0.00 / 0.00 &
        0.00 / 0.00 &
        2.67 / 2.67 \\

        sweep\_blocks (Rand.) &
        0.00 / 0.00 &
        0.00 / 0.00 &
        0.00 / 0.00 &
        0.00 / 0.00 &
        0.00 / 0.00 &
        1.33 / 1.33 \\

        pour\_liquid\_into\_cup (Std.) &
        0.00 / 0.00 &
        \underline{20.00} / \underline{20.00} &
        18.67 / 18.67 &
        \textbf{48.00} / \textbf{48.00} &
        28.00 / 28.00 &
        54.67 / 54.67 \\

        pour\_liquid\_into\_cup (Rand.) &
        0.00 / 0.00 &
        0.00 / 0.00 &
        0.00 / 0.00 &
        0.00 / 0.00 &
        1.33 / 1.33 &
        16.00 / 16.00 \\

        make\_toast (Std.) &
        2.00 / 0.00 &
        \textbf{15.00} / \textbf{4.00} &
        \underline{3.67} / 0.00 &
        \textbf{15.00} / \textbf{4.00} &
        9.33 / 1.33 &
        13.33 / 5.33 \\

        make\_toast (Rand.) &
        0.67 / 0.00 &
        \underline{1.00} / 0.00 &
        0.00 / 0.00 &
        \textbf{3.00} / 0.00 &
        2.67 / 0.00 &
        10.67 / 1.33 \\

        arrange\_largest\_number (Std.) &
        0.80 / 0.00 &
        \textbf{16.80} / \textbf{12.00} &
        1.20 / 0.00 &
        \underline{11.60} / \underline{4.00} &
        5.13 / 1.33 &
        7.67 / 1.33 \\

        arrange\_largest\_number (Rand.) &
        \textbf{0.27} / 0.00 &
        \underline{0.20} / 0.00 &
        0.13 / 0.00 &
        \underline{0.20} / 0.00 &
        0.40 / 0.00 &
        1.73 / 0.00 \\

        sort\_nesting\_dolls\_by\_size (Std.) &
        0.00 / 0.00 &
        \textbf{28.00} / \textbf{28.00} &
        1.33 / 1.33 &
        \underline{24.00} / \underline{24.00} &
        10.67 / 10.67 &
        14.67 / 14.67 \\

        sort\_nesting\_dolls\_by\_size (Rand.) &
        0.00 / 0.00 &
        0.00 / 0.00 &
        0.00 / 0.00 &
        0.00 / 0.00 &
        0.00 / 0.00 &
        0.00 / 0.00 \\

        store\_laptop\_and\_headphones (Std.) &
        0.80 / 0.00 &
        \underline{26.40} / \underline{8.00} &
        4.80 / 0.00 &
        \textbf{44.80} / \textbf{40.00} &
        19.20 / 9.33 &
        35.73 / 20.00 \\

        store\_laptop\_and\_headphones (Rand.) &
        0.00 / 0.00 &
        \underline{5.60} / 0.00 &
        0.27 / 0.00 &
        \textbf{6.40} / 0.00 &
        13.07 / 1.33 &
        20.80 / 2.67 \\

        stack\_blocks (Std.) &
        5.40 / 0.00 &
        \underline{22.00} / \underline{16.00} &
        8.33 / 5.33 &
        \textbf{40.20} / \textbf{36.00} &
        22.53 / 13.33 &
        49.67 / 42.67 \\

        stack\_blocks (Rand.) &
        0.00 / 0.00 &
        0.00 / 0.00 &
        0.00 / 0.00 &
        \textbf{0.60} / 0.00 &
        1.20 / 0.00 &
        3.40 / 0.00 \\

        \midrule
        \multicolumn{7}{l}{\textbf{Precision}} \\
        \midrule

        fasten\_screws &
        0.40 / 0.00 &
        \textbf{10.00} / 0.00 &
        1.60 / 0.00 &
        \underline{7.20} / 0.00 &
        15.13 / 2.67 &
        30.00 / 8.67 \\

        insert\_tubes &
        3.73 / 0.00 &
        \textbf{67.20} / \underline{54.00} &
        13.47 / 2.00 &
        \underline{66.80} / \textbf{56.00} &
        17.87 / 3.33 &
        58.53 / 42.67 \\

        plug\_in\_charger &
        0.00 / 0.00 &
        \underline{2.00} / \underline{2.00} &
        0.00 / 0.00 &
        \textbf{6.00} / \textbf{6.00} &
        0.00 / 0.00 &
        0.67 / 0.67 \\

        pour\_balls\_into\_vase &
        0.00 / 0.00 &
        \underline{16.00} / \underline{16.00} &
        5.33 / 5.33 &
        \textbf{24.00} / \textbf{24.00} &
        13.33 / 13.33 &
        28.00 / 28.00 \\

        play\_Xylophone &
        0.00 / 0.00 &
        0.00 / 0.00 &
        0.00 / 0.00 &
        0.00 / 0.00 &
        0.00 / 0.00 &
        0.00 / 0.00 \\

        deposit\_coin &
        2.80 / 0.00 &
        \textbf{13.60} / \textbf{6.00} &
        5.20 / \underline{1.33} &
        \underline{8.80} / \textbf{6.00} &
        3.20 / 0.67 &
        10.93 / 4.67 \\

        insert\_key &
        4.30 / 0.00 &
        \underline{13.50} / 0.00 &
        11.50 / 0.00 &
        \textbf{14.40} / 0.00 &
        11.90 / 0.00 &
        14.90 / 0.00 \\

        build\_tower &
        4.47 / 0.00 &
        34.60 / \underline{30.00} &
        \underline{42.07} / 26.00 &
        \textbf{68.40} / \textbf{60.00} &
        37.73 / 24.00 &
        82.93 / 78.67 \\

        \midrule
        \multicolumn{7}{l}{\textbf{Long-Horizon}} \\
        \midrule

        put\_bottles\_into\_dustbin &
        57.37 / 40.00 &
        \textbf{95.20} / \textbf{92.00} &
        48.90 / 28.00 &
        \underline{87.70} / \underline{82.00} &
        79.93 / 69.33 &
        96.30 / 94.00 \\

        play\_tic\_tac\_toe &
        0.00 / 0.00 &
        \textbf{5.80} / 0.00 &
        2.40 / 0.00 &
        \underline{5.20} / \textbf{2.00} &
        8.23 / 1.33 &
        65.23 / 40.00 \\

        classify\_objects &
        3.10 / \underline{1.33} &
        \underline{9.20} / \textbf{2.00} &
        2.90 / 0.00 &
        \textbf{9.30} / \textbf{2.00} &
        24.67 / 12.67 &
        10.33 / 4.00 \\

        fill\_pen\_holder &
        4.07 / 0.00 &
        \underline{40.60} / \underline{20.00} &
        17.57 / 6.00 &
        \textbf{44.20} / \textbf{24.00} &
        23.23 / 7.33 &
        41.27 / 18.67 \\

        fill\_egg\_holder &
        0.73 / 0.00 &
        \underline{2.20} / 0.00 &
        1.13 / 0.00 &
        \textbf{3.20} / 0.00 &
        2.23 / 0.00 &
        3.03 / 0.00 \\

        organize\_table &
        7.83 / 0.00 &
        \textbf{36.50} / \textbf{4.00} &
        22.33 / 0.00 &
        \underline{35.50} / \underline{2.00} &
        23.33 / 0.00 &
        46.33 / 11.33 \\

        play\_stacking\_toy &
        0.00 / 0.00 &
        0.00 / 0.00 &
        0.00 / 0.00 &
        0.00 / 0.00 &
        0.00 / 0.00 &
        0.47 / 0.00 \\

        make\_kong &
        0.00 / 0.00 &
        0.00 / 0.00 &
        \underline{18.00} / \underline{18.00} &
        \textbf{40.00} / \textbf{40.00} &
        26.67 / 26.67 &
        90.00 / 90.00 \\

        \midrule
        \multicolumn{7}{l}{\textbf{Memory}} \\
        \midrule

        cover\_blocks &
        0.00 / 0.00 &
        \underline{16.50} / \underline{10.00} &
        14.73 / \underline{10.00} &
        \textbf{18.90} / \textbf{12.00} &
        19.07 / 13.33 &
        20.67 / 14.67 \\

        match\_and\_pick\_from\_conveyor &
        20.67 / 20.67 &
        \textbf{36.00} / \textbf{36.00} &
        4.67 / 4.67 &
        \underline{26.00} / \underline{26.00} &
        14.00 / 14.00 &
        29.33 / 29.33 \\

        swap\_T &
        0.00 / 0.00 &
        0.00 / 0.00 &
        0.00 / 0.00 &
        0.00 / 0.00 &
        0.67 / 0.67 &
        0.00 / 0.00 \\

        press\_by\_number &
        0.00 / 0.00 &
        \textbf{2.00} / \textbf{2.00} &
        0.00 / 0.00 &
        0.00 / 0.00 &
        0.00 / 0.00 &
        0.00 / 0.00 \\

        imitate\_sorting\_sequence &
        \underline{0.63} / 0.00 &
        0.50 / 0.00 &
        \underline{0.63} / 0.00 &
        \textbf{1.00} / 0.00 &
        1.60 / 0.00 &
        1.67 / 0.00 \\

        swap\_blocks &
        0.00 / 0.00 &
        0.00 / 0.00 &
        0.00 / 0.00 &
        0.00 / 0.00 &
        0.00 / 0.00 &
        0.00 / 0.00 \\

        \midrule
        \multicolumn{7}{l}{\textbf{Open}} \\
        \midrule

        align\_blocks &
        0.00 / 0.00 &
        0.00 / 0.00 &
        0.00 / 0.00 &
        0.00 / 0.00 &
        0.00 / 0.00 &
        0.00 / 0.00 \\

        solve\_equation &
        0.00 / 0.00 &
        0.00 / 0.00 &
        0.00 / 0.00 &
        0.00 / 0.00 &
        0.00 / 0.00 &
        0.00 / 0.00 \\

        stack\_blocks\_by\_language &
        0.00 / 0.00 &
        0.00 / 0.00 &
        \underline{0.53} / 0.00 &
        \textbf{2.00} / \textbf{2.00} &
        1.73 / 0.00 &
        0.13 / 0.00 \\

        general\_pickup &
        3.33 / 3.33 &
        \textbf{14.00} / \textbf{14.00} &
        \underline{4.67} / \underline{4.67} &
        \textbf{14.00} / \textbf{14.00} &
        12.00 / 12.00 &
        12.67 / 12.67 \\

        classify\_objects\_by\_language &
        0.00 / 0.00 &
        \textbf{1.40} / 0.00 &
        0.20 / 0.00 &
        \underline{0.40} / 0.00 &
        0.60 / 0.00 &
        1.07 / 0.00 \\

        pick\_from\_conveyor\_by\_image &
        0.00 / 0.00 &
        0.00 / 0.00 &
        0.00 / 0.00 &
        0.00 / 0.00 &
        1.33 / 1.33 &
        0.00 / 0.00 \\

        store\_tools\_in\_toolbox &
        0.00 / 0.00 &
        0.00 / 0.00 &
        0.00 / 0.00 &
        0.00 / 0.00 &
        0.17 / 0.00 &
        0.00 / 0.00 \\

        pour\_by\_language &
        0.00 / 0.00 &
        0.00 / 0.00 &
        0.00 / 0.00 &
        0.00 / 0.00 &
        0.00 / 0.00 &
        0.00 / 0.00 \\

        \addlinespace[0.30em]
        \midrule
        \addlinespace[0.15em]

        \textbf{Average} &
        3.48 / 2.03 &
        \underline{13.33} / \underline{9.40} &
        6.40 / 3.24 &
        \textbf{15.56} / \textbf{11.70} &
        11.44 / 6.93 & 
        20.23 / 14.88 \\

        \addlinespace[0.15em]
        \bottomrule
    \end{tabular}
    }
\end{table*}

\section{Additional Results and Visualizations}
\label{app:additional}

\subsection{Complete BiCoord Results}
\label{app:bicoord}

Tab.~\ref{tab:bicoord_main} reports all 18 BiCoord tasks, complementing Tab.~\ref{tab:bicoord_LH} in the main text, which shows only the six with the longest expert trajectories.

\subsection{Complete RoboDojo Results}
\label{app:robodojo}
Tab.~\ref{tab:robodojo_taskwise} reports task-wise Score / Success Rate
for all 42 RoboDojo simulation tasks, complementing the capability-level
averages in Tab.~\ref{tab:robodojo_main}.
For the 12 Generalization tasks, the standard and randomized evaluation
settings are reported separately.

\subsection{Qualitative Rollouts}
\label{app:rollouts}

\begin{figure*}[t]
\centering
\includegraphics[width=\linewidth]{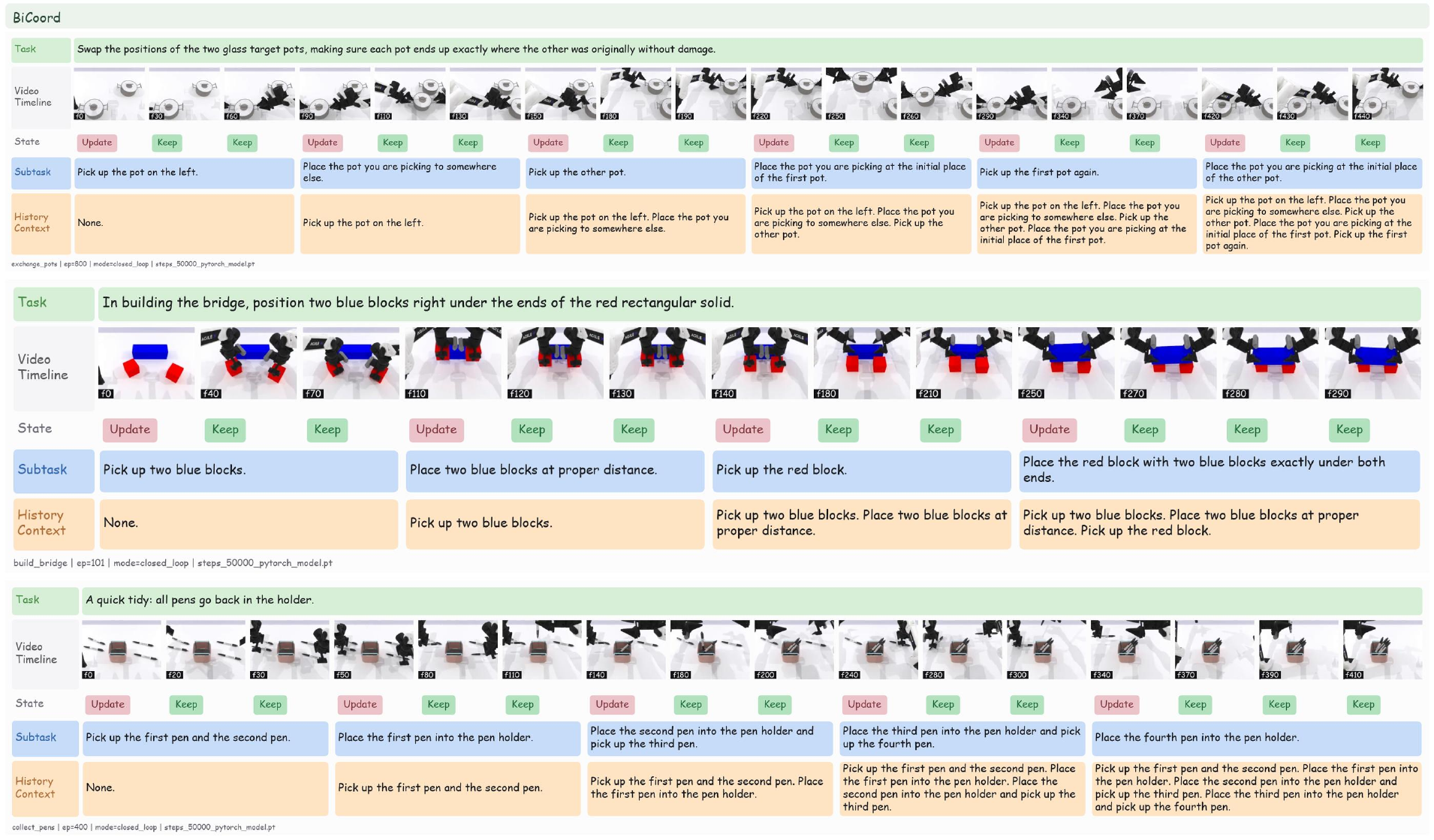}
\caption{
\textbf{Closed-loop rollouts on BiCoord.}
Bimanual tasks whose stages differ in which arm leads.
Subtask text carries the arm assignment, so a transition changes both what is being done and which arm does it.
}
\label{fig:sub_bicoord}
\end{figure*}

\begin{figure*}[h]
\centering
\includegraphics[width=\linewidth]{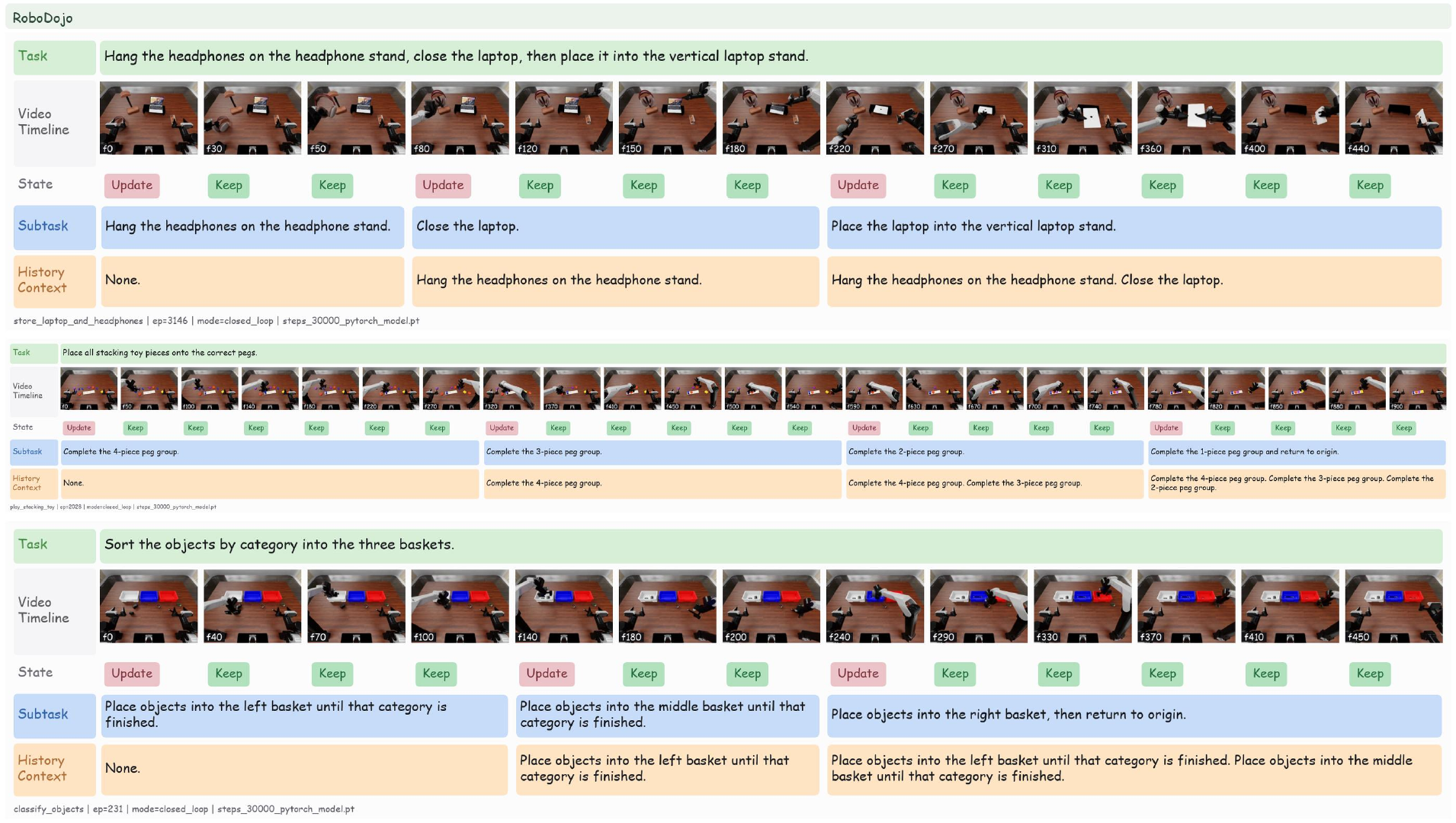}
\caption{
\textbf{Closed-loop rollouts on RoboDojo.}
Three episodes of increasing length.
The memory row grows one entry per \texttt{UPDATE}, so the subtask the policy is conditioned on is always paired with an explicit record of what preceded it.
}
\label{fig:sub_robodojo}
\end{figure*}

\begin{figure*}[h]
\centering
\includegraphics[width=\linewidth]{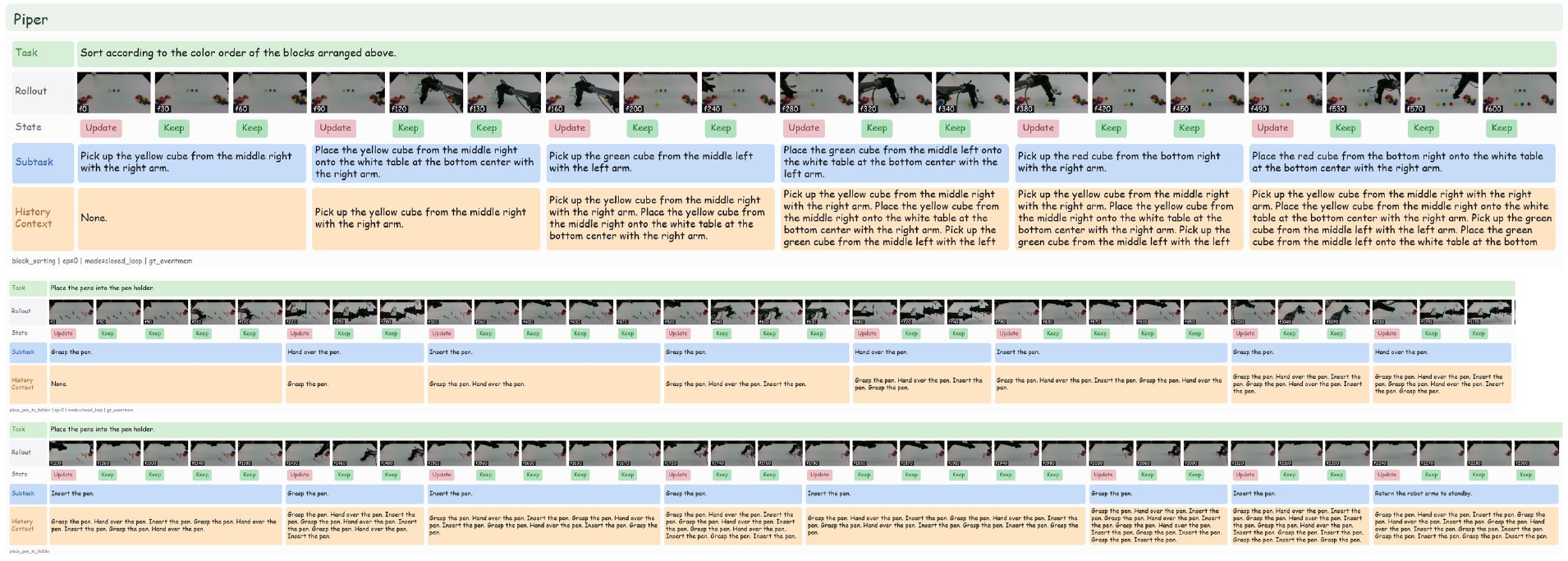}
\caption{
\textbf{Closed-loop rollouts on the real robot.}
\textit{Block Sorting} and \textit{Fill Pen Holder}.
The latter repeats one grasp--hand-over--insert cycle many times, so the memory is the only signal distinguishing the third repetition from the first. Zoom in for a better view.
}
\label{fig:sub_piper}
\end{figure*}

Figs.~\ref{fig:sub_bicoord}--\ref{fig:sub_piper} show closed-loop rollouts on the real robot, RoboDojo, and BiCoord. Each strip pairs the video timeline with the \texttt{KEEP}/\texttt{UPDATE} decision at every replanning step, the active subtask, and the accumulated memory. The pattern the strips make visible is the one Tab.~\ref{tab:update_strategy} quantifies: \texttt{UPDATE} fires a handful of times per episode, at subtask boundaries, and the state is simply carried in between.

\end{document}